\documentclass{article}

\usepackage[preprint]{corl_2026} 

\usepackage{graphicx}
\graphicspath{{images/}}

\usepackage[utf8]{inputenc} 
\usepackage[T1]{fontenc}    
\usepackage{hyperref}       
\usepackage{url}            
\usepackage{booktabs}       
\usepackage{amsfonts}       
\usepackage{nicefrac}       
\usepackage{microtype}      
\usepackage{xcolor}         

\usepackage{hyperref}
\usepackage{url}

\usepackage[utf8]{inputenc} 
\usepackage[T1]{fontenc}    
\usepackage{hyperref}       
\usepackage{url}            
\usepackage{booktabs}       
\usepackage{amsfonts}       
\usepackage{nicefrac}       
\usepackage{microtype}      
\usepackage{xcolor}         
\usepackage{enumitem}

\usepackage{pifont,xcolor}
\usepackage[legacycolonsymbols]{mathtools}
\usepackage{xcolor,colortbl}
\definecolor{LightRed}{HTML}{FFE0E0}
\definecolor{LightGray}{gray}{0.9} 
\usepackage{caption}  
\usepackage{booktabs} 
\definecolor{LightGreen}{rgb}{0.85, 0.95, 0.85}  
\usepackage{multirow}  
\usepackage{comment}
\usepackage{wrapfig}   

\usepackage{mdframed}
\usepackage{fancyvrb}
\definecolor{PromptGray}{gray}{0.95}
\newmdenv[
  backgroundcolor=PromptGray,
  linewidth=0pt,
  innerleftmargin=8pt,
  innerrightmargin=8pt,
  innertopmargin=6pt,
  innerbottommargin=6pt,
  skipabove=6pt,
  skipbelow=6pt,
]{promptbox}

\newcommand{\tbd}[1]{--}

\title{VLM-Guided Contact Retargeting and Generative Partner Modeling for Humanoid-Human Interaction}
\author{
  Yuto Shibata$^{1,2,3*}$\quad
  Kotaro Amaya$^{1,2,3*}$\quad
  Kashu Yamazaki$^{1}$\quad
  Lalit Jayanti$^{1}$\\
  \bfseries Yoshimitsu Aoki$^{2,3}$\quad
  Mariko Isogawa$^{2,3}$\quad
  Katerina Fragkiadaki$^{1}$\\[4pt]
  {\normalfont $^{1}$Carnegie Mellon University\quad
  $^{2}$Keio AI Research Center\quad
  $^{3}$Keio University}\\
  {\normalfont\small $^{*}$Equal contribution}\\[2pt]
  {\normalfont\small \url{https://yutoshibata07.github.io/REACT/}}
}

\begin{document}
\maketitle

\begin{center}
    \includegraphics[width=1.0\linewidth]{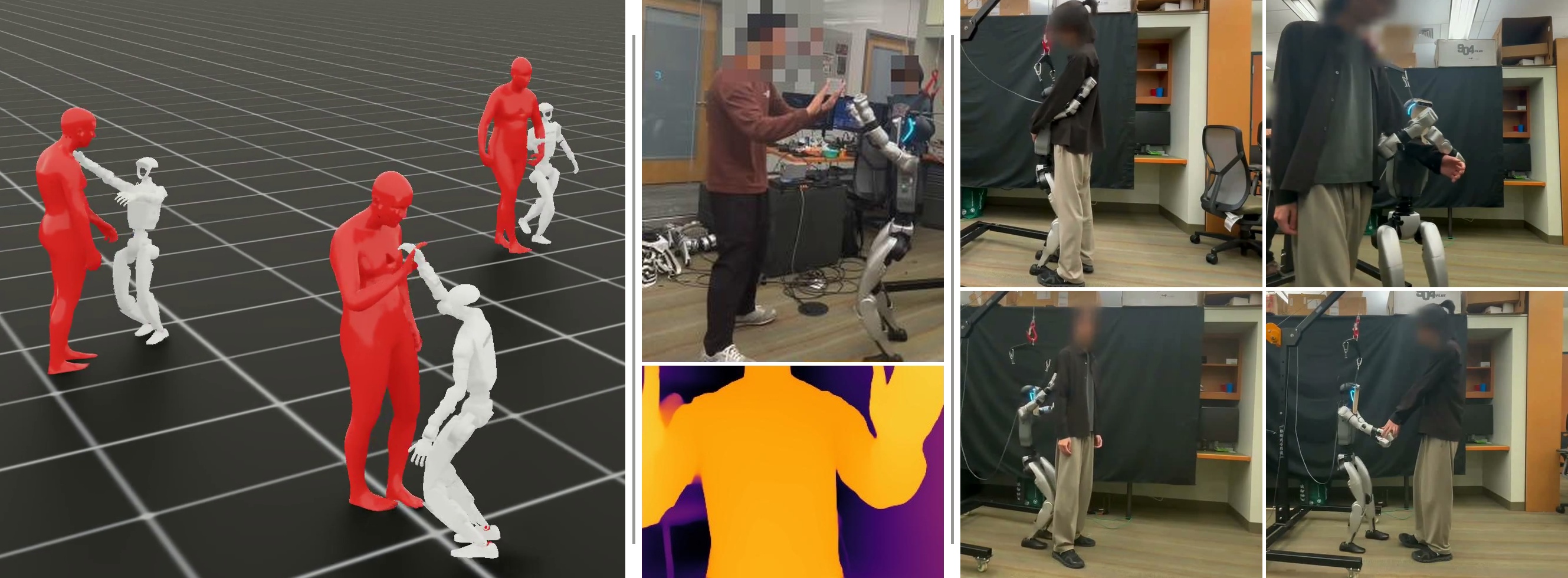}
    \captionof{figure}
    {
        Trained in simulation with contact-preserving retargeting and human depth rendering (left), our policy runs zero-shot on a real Unitree G1 (middle: high-five with depth input; right: others).
    }
    \label{fig:teaser}
\end{center}

\begin{abstract}
Humanoid–human interaction requires robots to operate under severe partial observability: during close physical contact, the human partner is occluded, truncated, or outside the robot’s field of view, causing pose estimation to fail and often necessitating external tracking or privileged human state. At the same time, heavy mutual occlusions introduce errors into the human–human motion-capture data used for learning. We present REACT, a framework for learning deployable humanoid–human interaction policies from human–human demonstrations. REACT treats semantic contact structure as the key supervisory signal connecting demonstration processing, interaction learning, and onboard control. First, a VLM-guided contact modeling and retargeting pipeline identifies semantically meaningful contacts and transfers them onto humanoid–human geometry while preserving interaction structure. Using these annotations, privileged interaction policies learn physically grounded behaviors through interaction-aware contact rewards and are then distilled into deployable egocentric depth policies trained entirely from onboard sensing using photorealistic human depth rendering. To address severe partial observability, we introduce generative partner-state modeling objectives that predict partner motion from occluded egocentric views. We evaluate REACT in simulation and on a real Unitree G1 humanoid across five contact-rich interactions. Semantic contact modeling substantially improves preservation of demonstrated interactions, while generative partner-state modeling improves contact recall on four of five interactions, most under heavy occlusion. On real hardware, REACT performs all five interactions using onboard sensing alone.
\end{abstract}


\section{Introduction}
\label{sec:introduction}

Humanoid hardware and learning-based control have advanced rapidly in recent years, bringing the deployment of humanoid robots in homes, workplaces, and care settings increasingly within reach. Realizing this vision, however, requires robots capable of sustained, reactive physical interaction with humans using only onboard sensing. Unlike existing humanoid controllers, which primarily focus on isolated motion tracking or interaction with static scenes and objects~\cite{he2024omnih2o,ji2026exbody2,ben2025homie}, humanoid--human interaction (HHI) requires continuous perception, prediction, and physical coordination with a human partner whose motion evolves under independent intent and frequent contact.

A central challenge in HHI is that close-contact interaction fundamentally operates under severe partial observability. During interaction, the human partner is frequently self-occluded, truncated, or partially outside the robot's field of view, causing off-the-shelf human pose-estimation systems to fail precisely when accurate perception matters most. As a result, many existing approaches rely on external tracking systems, motion-capture instrumentation, or privileged access to human state during deployment, limiting robustness and real-world deployability. The problem begins even earlier in the learning pipeline: heavy mutual occlusions also corrupt the underlying human--human motion-capture data itself, introducing reconstruction artifacts and inconsistent contact geometry that propagate through downstream retargeting~\cite{joao2025gmr,yang2025omniretarget} and policy-learning pipelines.

We argue that deployable humanoid--human interaction fundamentally requires recovering the interaction state from incomplete observations, and that semantic contact structure provides the key supervisory signal connecting noisy demonstrations, physics-based learning, and onboard control. While motion trajectories may vary substantially across embodiments, interactions are often defined by a much more stable structure: who touches whom, where, and when. We therefore propose to represent interactions through semantic contacts and train policies to infer latent partner state under severe partial observability.

To this end, we introduce REACT (\textbf{R}eactive \textbf{E}gocentric \textbf{A}gent for \textbf{C}ontact-rich Humanoid-Human In\textbf{T}eraction), a framework for learning deployable humanoid--human interaction policies from large-scale human--human motion-capture data (Figure~\ref{fig:teaser}). REACT first runs a VLM-guided semantic contact pipeline that retargets the demonstrated interaction contacts onto humanoid--human geometry while preserving interaction structure across the embodiment gap. These semantic contacts then supervise a privileged interaction teacher through interaction-aware contact rewards, anchoring the policy in physics. The teacher is distilled into an egocentric depth student trained entirely on onboard sensing via photorealistic in-simulator SMPL-X rendering of dynamic human partners. To address severe partial observability, the student is further trained with a generative partner-state prediction objective that infers past, present, and future partner motion from egocentric observations.

We evaluate REACT on a large corpus of close-contact human--human interactions in simulation and on a real Unitree G1 humanoid. Our experiments show that semantic contact modeling substantially improves preservation of demonstrated interaction structure compared to existing retargeting approaches, and its contact annotations improve downstream interaction learning. We further show that generative partner-state prediction improves contact recall on four of five interactions, most where the partner is hardest to see, enabling depth-based policies to recover most of the contact recall of privileged interaction teachers ($75\%$ at a $30\,$cm tolerance). On real hardware, REACT successfully executes diverse contact-rich interactions, including high-fives, hugs, handshakes, link-arms, and pats on the back, using onboard sensing alone.

In summary, our contributions are 
(1) A semantic contact modeling and retargeting framework that transfers human--human demonstrations to humanoid--human interactions while preserving interaction structure across embodiments, 
(2) An interaction-learning framework that uses semantic contacts as supervision for physics-based humanoid--human behavior learning, 
(3) Generative partner-state prediction objectives that enable latent interaction-state inference under severe partial observability from egocentric observations, and 
(4) extensive simulation ablations and real-world deployment on a Unitree G1 across diverse contact-rich humanoid--human interactions.

\section{Related Work}
\label{sec:related-work}

\paragraph{Humanoid motion tracking and scene interaction}
Recent work has achieved impressive progress in humanoid teleoperation and motion tracking through imitation learning, teacher--student distillation, and large-scale behavioral modeling~\cite{he2024omnih2o,he2024hover,ji2026exbody2,he2025asap,chen2025gmt,zhang2025any2track,luo2025sonic,zeng2025bfm,li2026bfmzero}. A growing body of work further extends humanoid control toward object and scene interaction through manipulation and loco-manipulation, physics-based human--object interaction, scene-conditioned motion generation, and visual sim-to-real learning~\cite{zhao2025resmimic,weng2025hdmi,qiu2025humanoidpolicy,li2025thor,xu2025intermimic,wang2025physhsi,yin2025visualmimic,he2025viral,han2025opendoor,kalaria2026dreamcontrol,videomimic}. However, these methods either treat the humanoid in isolation or restrict interaction to static scenes and inanimate objects, whereas REACT targets closed-loop humanoid--human interaction with a dynamically moving human partner.

\paragraph{Interaction Modeling and Human--Human Motion Synthesis}
Human--human interaction synthesis has advanced rapidly through diffusion-based and generative motion models. 
Recent methods such as InterMask~\cite{javed2025intermask}, PINO~\cite{ota2025pino}, UMO~\cite{cong2026umo}, and MAGNet~\cite{maluleke2025magnet} further improve motion diversity, long-horizon interaction synthesis, and flexible multi-agent conditioning. However, these approaches primarily operate in kinematic space and therefore lack physical grounding and closed-loop interaction control. More recent physics-based interaction methods couple diffusion-based motion generation with physics-based tracking~\cite{ji2025humanx,shibata2026assistmimic} or forward-dynamics-guided imitation~\cite{liu2024physreaction}, but they generate the reference interaction offline~\cite{shibata2026assistmimic} or observe the partner fully, either in simulation~\cite{liu2024physreaction} or through an unoccluded third-person capture~\cite{ji2025humanx}, rather than through deployable onboard sensing.

\paragraph{Interaction retargeting}
Motion retargeting offers a promising avenue for transferring large-scale human interaction data to humanoid robots. Optimization-based methods, ranging from inverse kinematics to trajectory optimization~\cite{choi2019naturalmotion,gomes2019movement,joao2025gmr}, adapt the source motion to the humanoid's kinematics frame by frame or over the whole trajectory, while learning-based retargeting~\cite{choi2021selfsupervised} improves stylistic consistency and motion realism. However, because their objectives are defined on each body separately, they carry no notion of which body parts are meant to touch the partner, so contacts present in the source interaction are easily lost once the motion is transferred to a different embodiment. Graphics-oriented approaches have explored geometry-aware and contact-aware retargeting objectives~\cite{cheynel2025reconform,jang2024geometryaware}, while recent interaction-preserving methods~\cite{yang2025omniretarget} emphasize maintaining relational geometry for downstream policy learning. Beyond Mimicry~\cite{huang2026beyond} retargets human--human interactions to a humanoid by preserving pairwise keypoint distances between the two agents, without reasoning about which of them correspond to physical contact. In contrast, REACT introduces semantic contact-aware retargeting that explicitly preserves interaction contacts and relational geometry while suppressing interpenetration between the humanoid and its partner.

\paragraph{Interaction under Partial Observability}
Closer to our setting, recent work has approached human--robot interaction by learning from human--human--object demonstrations~\cite{chen2025rhino}, forecasting social motion to generate robot behavior~\cite{mascaro2024echo}, and adding compliance to whole-body control for physical contact with people~\cite{lu2025gentlehumanoid}. 
However, these systems depend on robot-side teleoperation data~\cite{chen2025rhino}, external motion capture~\cite{lu2025gentlehumanoid}, or a separate human-pose estimator~\cite{huang2026beyond,ji2025humanx}, which becomes unreliable at the close range where physical contact occurs.
Privileged-to-perceptual distillation has produced visual humanoid policies~\cite{he2025viral,han2025opendoor,yin2025visualmimic}, but only for interaction with objects and scenes rather than a moving human partner. In contrast, REACT maps egocentric depth to actions end-to-end and explicitly trains policies to infer hidden 
partner motion under occlusion through generative 
partner-state modeling.

\section{Method}
\label{sec:method}

\begin{figure}[t]
\centering
\centerline{\includegraphics[width=1.0\linewidth]{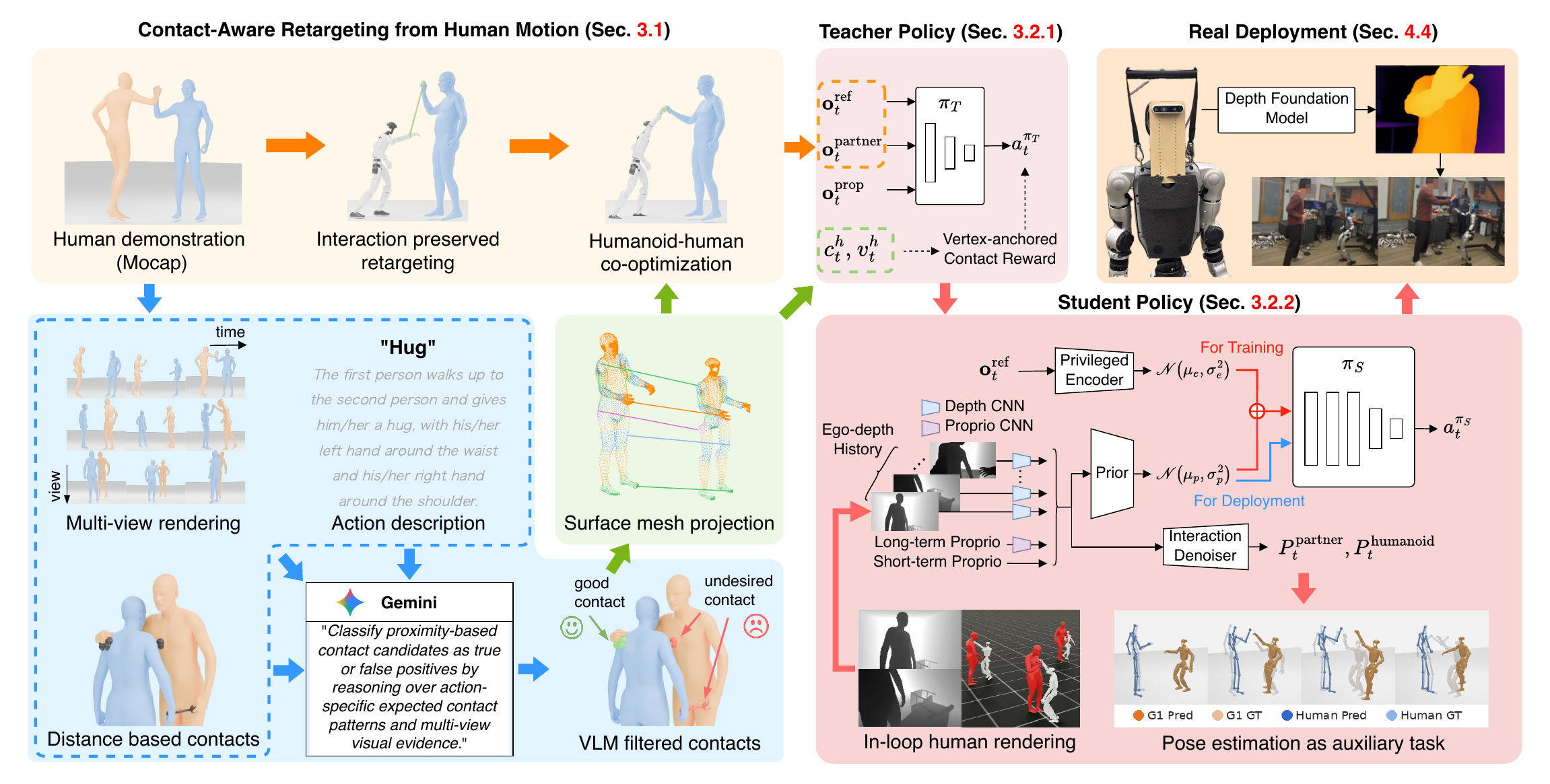}}
\vspace{-2mm}
\caption{\textbf{Overview of REACT.} REACT first annotates semantic contacts in human--human demonstrations with a VLM and retargets them onto the humanoid while preserving them. A privileged teacher then learns the interaction with contact rewards and is finally distilled into an egocentric depth student that predicts the partner's motion and runs on the real robot.}
\label{fig:pipeline}
\vspace{-4mm}
\end{figure}

We formulate deployable humanoid–human interaction as recovering and preserving interaction state across three stages: (i) inferring semantic contact structure from noisy human–human demonstrations, (ii) learning physically grounded interaction policies that reproduce this contact structure under physics, and (iii) inferring partner state from egocentric observations during deployment.

Figure~\ref{fig:pipeline} provides an overview of REACT. Given a human--human motion-capture sequence, REACT learns a humanoid policy that reproduces the demonstrated interaction from onboard sensing alone, in three stages. First, \emph{semantic contact modeling} identifies meaningful contacts and transfers them across the embodiment gap into humanoid--human references. Second, a privileged \emph{teacher} learns the interaction with contact rewards derived from these annotations. Finally, a deployable \emph{egocentric student} is distilled from the teacher using only onboard depth and infers partner state through generative prediction.

\subsection{Semantic Contact Modelling and Retargeting}
\label{sec:method:contact}
Existing retargeting methods optimize pose similarity or relational geometry but do not explicitly reason about the contact structure that defines an interaction~\cite{joao2025gmr,yang2025omniretarget}. REACT instead represents an interaction as a collection of spatiotemporal semantic contacts and preserves them throughout embodiment transfer. Semantic contact modeling proceeds in three steps, detailed below:

\noindent
\textbf{Human-to-humanoid surface correspondence}\quad 
Semantic contacts are expressed on SMPL-X body vertices, but the retargeted humanoid has a different morphology with no direct vertex correspondence. We therefore register the SMPL-X mesh to the humanoid surface using Laplacian-regularized non-rigid ICP (Fig.~\ref{fig:pipeline}, \emph{Surface mesh projection}), producing an offline, pose-independent mapping from every SMPL-X vertex to a corresponding humanoid surface location $(\text{link}, \mathbf{p}_{\text{local}})$. The resulting correspondence transfers semantic contacts from the human body onto the humanoid embodiment and provides the geometric reference used by both retargeting and interaction-reward computation.

\noindent
\textbf{VLM-guided semantic contact annotation}\quad 
A key challenge is that existing motion-capture datasets do not provide annotations of \emph{who touches whom, where, and when}. Furthermore, reconstructed human meshes are often noisy under heavy mutual occlusion, making geometric proximity alone an unreliable indicator of interaction contact. Simple distance thresholds either miss meaningful contacts or produce large numbers of spurious contact events.

We therefore separate geometric contact discovery from semantic verification. A permissive mesh-distance pass first proposes candidate spatiotemporal contacts between the two interacting humans. These candidates are subsequently verified by a vision-language model (Gemini~2.0 Flash), which receives multi-view renderings of the interaction, the candidate contact list, and a description of the interaction type depicted in the video. The VLM rejects contacts inconsistent with the semantics of the demonstrated interaction and retains only meaningful contact events (Fig.~\ref{fig:pipeline}, \emph{Distance based contacts} to \emph{VLM filtered contacts}). For each contact site $h$ at timestep $t$, the pipeline emits a flag $c_t^h$, the actor vertex $v_t^h$, and the contacted partner vertex $j_t^h$ used downstream by retargeting and the contact reward. Implementation details and the annotation-quality evaluation are in Appendix~\ref{sec:supp:contact-annotation}, and the VLM prompt is in Appendix~\ref{sec:supp:vlm-prompt}.

\noindent
\textbf{Semantic contact retargeting}\quad 
Given surface correspondences and semantic contact annotations, we retarget human–human demonstrations into humanoid–human interaction references while explicitly preserving demonstrated contact structure across the embodiment gap.

\emph{Stage 1: Interaction-aware kinematic retargeting.}
We begin with an OmniRetarget-style initialization~\cite{yang2025omniretarget} (Fig.~\ref{fig:pipeline}, \emph{Interaction preserved retargeting}). Only the human mapped to the G1 is rescaled to the G1 embodiment, because rescaling the partner would move the contact targets it supplies and destabilize the G1 retarget. This breaks the inter-body interaction geometry across the embodiment gap, which we mitigate by anchoring the humanoid arm end-effector targets to the demonstrated contact locations on labeled frames.

\emph{Stage 2: Semantic contact driven human-humanoid co-optimization.}
The initialization is subsequently refined through a contact-preserving optimization stage (Fig.~\ref{fig:pipeline}, \emph{Humanoid-human co-optimization}). For every annotated contact $(c_t^h = 1)$, the corresponding humanoid contact point is driven toward the target partner vertex $j_t^h$. At the same time, two complementary regularizers preserve the remainder of the interaction geometry. A distance-preserving proximity field maintains the demonstrated separation between non-contact body regions, preventing collapse of the interaction configuration, while a volumetric SDF penalty~\cite{ICCV25:VolumetricSMPL} suppresses interpenetration. Optimization alternates between humanoid and partner kinematics while preserving the locomotion generated during Stage~1.

The resulting trajectories preserve both the motion and the semantic contact structure of the original interaction. These retargeted demonstrations provide the reference motions used for policy learning and the semantic contact annotations used by the interaction rewards described in Sec.~\ref{sec:method:contact-reward}.

\subsection{Interaction Learning from Semantic Contacts}
\label{sec:method:teacher} 
We use the semantic contact annotations of Sec.~\ref{sec:method:contact} to learn a privileged interaction teacher that reproduces the demonstrated contact structure under physics.

\subsubsection{Privileged Interaction Teacher}
\label{sec:method:teacher_policy} 

We train a privileged policy $\pi_T$ using PPO~\cite{schulman2017proximal} (Fig.~\ref{fig:pipeline}, \emph{Teacher Policy}). The teacher has access to the full state of the human partner and therefore serves as an upper bound on interaction performance before addressing the challenges of onboard perception. 
The teacher receives three groups of observations: (i) proprioceptive robot state $\mathbf{o}^{\text{prop}}_t$, including base motion, joint states, end-effector positions, and previous actions; (ii) reference-motion observations $\mathbf{o}^{\text{ref}}_t$ from the retargeted demonstrations; and (iii) privileged partner-state observations $\mathbf{o}^{\text{partner}}_t$ containing the partner's pose and motion expressed in the humanoid's egocentric frame. More observation details are provided in Appendix~\ref{sec:supp:teacher}.

Rather than directly controlling all robot joints, the teacher predicts a hybrid action consisting of locomotion commands and upper-body interaction actions. Specifically, we leverage the pretrained HOMIE controller\cite{ben2025homie} for stable locomotion and predict only four locomotion commands $(v_x,v_y,\omega_z,h)$ together with residual upper-body joint offsets. This factorization allows the policy to focus its capacity on learning contact-rich interaction while inheriting robust locomotion capabilities.

\paragraph{Interaction-Aware Contact Rewards}
\label{sec:method:contact-reward} 

Tracking the reference alone does not guarantee reproducing the
interaction: a policy can match the reference trajectory while still
failing to establish or maintain contact. We therefore supervise the
teacher with the semantic contact annotations of
Sec.~\ref{sec:method:contact} (Fig.~\ref{fig:pipeline}, \emph{Vertex-anchored Contact Reward}). Prior contact-based rewards rely on
hand-annotated or geometric-proximity contacts~\cite{shibata2026assistmimic,ji2025humanx};
ours instead consume the automatically extracted semantic contacts.
The analytic form of the rewards is in Appendix~\ref{sec:supp:contact-reward}.

\subsubsection{Deployable Perceptive Student}
\label{sec:method:student}

The privileged interaction teacher assumes among others access to the full state of the human partner and therefore cannot be deployed on a real humanoid. In practice, close-contact interaction must operate from onboard sensing alone, where severe occlusions, truncation, and limited field of view make the partner's state only partially observable. We therefore distill the privileged teacher into a deployable egocentric student policy $\pi_S$ (Fig.~\ref{fig:pipeline}, \emph{Student Policy}) that learns to infer interaction state directly from an onboard depth stream and proprioception history. Depth input images are processed by a shared ResNet encoder~\cite{He_2016_CVPR}, while a temporal convolutional network aggregates long-horizon proprioceptive history into a compact motion-context representation. The resulting deployable observation embedding $\mathbf{c}_t$ is passed to a conditional variational autoencoder. The student policy is trained using on-policy DAgger~\cite{ross2011dagger}.

\paragraph{Photorealistic Human Rendering}
\label{sec:method:rendering}
Most physics simulators represent humans with simplified collision
primitives such as capsule colliders~\cite{luo2023perpetual},
producing depth observations that diverge sharply from real human
depth. We close this gap with an in-loop rendering pipeline (Fig.~\ref{fig:pipeline}, \emph{In-loop human rendering}) that
attaches a visual SMPL-X mesh to the partner: the capsule collider is
kept for physics, while a batched SMPL-X forward pass updates the
visual mesh every control step so the robot's depth camera sees
realistic human geometry, which closes the sim-to-real gap (Fig.~\ref{fig:depth-compare} in Appendix~\ref{sec:supp:rendering}).



\paragraph{Generative Partner-State Modeling}
\label{sec:method:partner_model}
We additionally train the student to infer the hidden state of the interaction.
In humanoid--human interaction, the partner is rarely fully observable: during close contact, large portions of their body are frequently occluded or outside the camera field of view. Consequently, many different partner configurations may be consistent with the same visual observation.
We therefore train the student to predict how the interaction evolves from these partial observations: a generative partner-state prediction objective predicts the partner's past, present, and future motion from the deployable observation representation $\mathbf{c}_t$. A diffusion-based prediction head (Fig.~\ref{fig:pipeline}, \emph{Interaction Denoiser}) implements this objective, outputting the partner's SMPL-X body pose, joint velocities, and orientations, $P^{\text{partner}}_t$, together with the humanoid's short-horizon future trajectory, $P^{\text{humanoid}}_t$ (time offsets in Appendix~\ref{sec:supp:student}). These auxiliary predictions provide dense supervision for the visual encoder and encourage the learned latent representation to capture interaction-relevant information that is not directly observable from a single depth frame. Rather than reacting only to visible motion, the student learns to anticipate future interaction dynamics.

\paragraph{Training objective} The final training objective combines teacher-action imitation, latent-state distillation, and generative partner-state prediction:
\begin{equation}
\mathcal{L}
=
\mathcal{L}_{\text{BC}}
+
\lambda_{\text{KL}}
\mathcal{L}_{\text{KL}}
+
\lambda_{\text{gen}}
\mathcal{L}_{\text{gen}}
+
\lambda_{\text{pose}}
\mathcal{L}_{\text{pose}},
\label{eq:student-loss}
\end{equation}
where the individual loss terms correspond to behavior cloning, latent-state alignment, generative partner-state prediction, and single-step pose estimation, respectively. By jointly learning interaction control and inference of the occluded partner's state, the resulting policy remains robust under severe occlusion and partial observability while operating entirely from onboard sensing.

\section{Experiments}
\label{sec:experiments}

We evaluate REACT on both simulation and real-world humanoid–human interaction tasks. Our experiments are designed to answer five questions:

\begin{itemize}
\item[\textbf{Q1.}] Does VLM-guided contact annotation improve interaction retargeting quality?
\item[\textbf{Q2.}] Do semantic contact annotations improve downstream interaction policy learning?
\item[\textbf{Q3.}] Does generative partner-state modeling improve robustness under partial observability?
\item[\textbf{Q4.}] Can an egocentric depth policy recover the performance of a privileged interaction teacher?
\item[\textbf{Q5.}] Does REACT transfer successfully to real humanoid hardware?
\end{itemize}

\noindent
\textbf{Dataset and simulation setup}\quad 
We train on five interactive Inter-X~\cite{interx} actions (high-five, hug, pat-on-back, handshake, link arms) with a manually filtered $9{:}1$ train/evaluation split. All policies are trained in Isaac Sim with the human partner replayed via PD control; the teacher receives privileged partner kinematics, while the student observes only onboard egocentric depth. Full data statistics and the filtering procedure are in Appendix~\ref{sec:supp:data}, and the rendering setup is in Appendix~\ref{sec:supp:rendering}.

\noindent\textbf{Metrics}\quad
For retargeted motion we report \emph{Jitter} and \emph{Penetration}.
Contact correctness in the VLM annotation and retargeting stages is
measured with \emph{Recall}, \emph{Precision}, and
\emph{$F_1$} against $80$ carefully hand-annotated GT contact clips;
the teacher and student policies report \emph{Contact Recall} against the
VLM-verified contact annotations (manual annotation does not scale to these
pools), the teacher on all motions of the student pool (training and held-out) at $15\,$cm and the student on the
held-out split at $30\,$cm, and the teacher additionally reports \emph{MPJPE} against the retargeted reference. Perceived quality uses a \emph{user study} (mean
rank, lower is better, $N{=}28$ participants). Metric definitions, thresholds, and GT sources are in
Appendices~\ref{sec:supp:retarget-eval} and~\ref{sec:supp:teacher}, and the user-study protocol is in Appendix~\ref{sec:supp:userstudy}.

\subsection{Evaluating Semantic-Aware Contact Modelling and  Retargeting}
\label{sec:exp:retarget}

We first ask whether REACT's retargeting stage transfers the \emph{interaction},
not merely the \emph{motion}, of a human--human demonstration onto the
humanoid--partner pair. We compare against GMR~\cite{joao2025gmr}, a kinematic
whole-body retargeting baseline, and OmniRetarget~\cite{yang2025omniretarget}, a
contact-and-collision-aware retargeting method. \emph{Ours w/o VLM contact
annotation} ablates the per-frame contact labels and target joints
(Sec.~\ref{sec:method:contact-reward}), falling back to the
distance-based contact candidates from Stage~1 of our pipeline
without the VLM verification step.

\begin{table*}[t]
  \centering
  \caption{\textbf{Contact-aware retargeting}}
  \vspace{-2mm}
  \label{tab:retarget}
  \resizebox{\textwidth}{!}{
  \begin{tabular}{lccccc}
    \toprule
    Method & Jitter\,$\downarrow$ & Penetration\,$\downarrow$ & Recall\,$\uparrow$ & Precision\,$\uparrow$ & User Study\,$\downarrow$ \\
           & (m/s$^3$) & (cm) & (\%) & (\%) & (rank 1--4) \\
    \midrule
    GMR                                 & $\mathbf{134.1}_{\pm 96}$ & $2.14_{\pm 3.05}$ & $0.42_{\pm 3.70}$ & $2.26_{\pm 9.72}$ & $3.61_{\pm 0.04}$ \\
    OmniRetarget                        & $560.4_{\pm 341}$ & $1.31_{\pm 1.85}$ & $22.13_{\pm 34.24}$ & $12.51_{\pm 18.47}$ & $2.57_{\pm 0.04}$ \\
    \midrule
    Ours w/o Stage 2 co-opt & $153.7_{\pm 71}$ & $1.31_{\pm 1.73}$ & $13.39_{\pm 26.56}$ & $11.79_{\pm 19.58}$ & --- \\
    Ours w/o VLM filter & $161.7_{\pm 76}$ & $1.04_{\pm 1.11}$ & $58.02_{\pm 41.71}$ & $11.60_{\pm 14.39}$ & $2.06_{\pm 0.03}$ \\
    \rowcolor{LightGreen}
    Ours                                & $155.4_{\pm 71}$ & $\mathbf{0.96}_{\pm 1.10}$ & $\mathbf{59.43}_{\pm 38.91}$ & $\mathbf{14.17}_{\pm 15.46}$ & $\mathbf{1.76}_{\pm 0.03}$ \\
    \bottomrule
  \end{tabular}
  }
\vspace{-4mm}
\end{table*}

\begin{table*}[t]
  \centering
  \caption{\textbf{Teacher tracking results.} Contact recall at 15\,cm, mean $\pm$ std over 3 seeds.}
  \vspace{-2mm}
  \resizebox{\linewidth}{!}{
  \label{tab:teacher}
  \begin{tabular}{l cc cc cc}
    \toprule
    & \multicolumn{2}{c}{High-five} & \multicolumn{2}{c}{Hug} & \multicolumn{2}{c}{Link-arms} \\
    \cmidrule(lr){2-3} \cmidrule(lr){4-5} \cmidrule(lr){6-7}
    Method & MPJPE\,$\downarrow$ & Contact\,$\uparrow$ & MPJPE\,$\downarrow$ & Contact\,$\uparrow$ & MPJPE\,$\downarrow$ & Contact\,$\uparrow$ \\
           & (cm) & (\%) & (cm) & (\%) & (cm) & (\%) \\
    \midrule
    \rowcolor{LightGreen}
    Ours                      & \textbf{6.5} & $\mathbf{70.9 \pm 0.3}$ & 8.3 & $\mathbf{53.1 \pm 0.5}$ & \textbf{7.2} & $\mathbf{74.8 \pm 0.1}$ \\
    (-) mesh contact reward   & \textbf{6.5} & $39.5 \pm 0.3$ & 9.5 & $31.4 \pm 0.1$ & 8.6 & $40.5 \pm 0.4$ \\
    (-) VLM contact filtering & 6.6 & $67.7 \pm 0.2$ & \textbf{8.1} & $46.1 \pm 0.2$ & 7.7 & $73.4 \pm 0.4$ \\
    \bottomrule
  \end{tabular}
  }
  \vspace{-4mm}
\end{table*}

Table~\ref{tab:retarget} reports the four geometry metrics together
with the user study; pool definitions and contact-event GT are in
Appendix~\ref{sec:supp:retarget-eval}, and the user-study protocol is in Appendix~\ref{sec:supp:userstudy}.

\noindent\textbf{Contact-preserving retargeting (Q1)}\quad The baselines mark opposite failure modes of motion-only
retargeting: GMR is smooth but contact-blind, while OmniRetarget is
collision-aware at a large jitter cost and its hard collision
constraints leave $\approx\!30\%$ of clips infeasible. Our \emph{Stage~1} ablation matches
OmniRetarget on Penetration and Precision while reducing jitter to GMR's level, inheriting its contact-aware geometry. Stage~2
co-optimisation then jumps Recall from $13.4\%$ to $58.0\%$ at
unchanged jitter, and the VLM verifier further lifts Precision from
$11.6\%$ to $14.2\%$ and Recall to $59.4\%$, matched by a $0.30$-rank
user-study improvement on overall preference.

\subsection{Teacher Policy: Tracking the Interaction}
\label{sec:exp:teacher}

We next evaluate the privileged-state teacher (Sec.~\ref{sec:method:teacher}), which must reproduce the retargeted interaction under physics. We ablate the two components that distinguish it from a standard tracking controller. The first is the \textbf{mesh contact reward}, which utilizes the ICP-based mesh-level contact positions and directions of humanoid and SMPL. In the \emph{(-) mesh contact reward} baseline, it is replaced by a link-level reward that pulls the center of each G1 contact link (the palm for the hands) toward the annotated partner joint, without the mesh registration or the directional and force-band terms. In the \emph{Ours w/o VLM contact filtering} baseline, the contact positions and timings used for this reward are instead calculated purely based on distance.

\noindent\textbf{Contact supervision for policy learning (Q2)}\quad Table~\ref{tab:teacher} demonstrates that both contact-related components contribute on the most contact-rich actions: the mesh contact reward lifts Link-arms Contact Recall by $+34.3\,$pt and Hug by $+21.7\,$pt, while VLM contact filtering lifts Hug by $+7.0\,$pt, High-five by $+3.2\,$pt, and Link-arms by $+1.4\,$pt, with the two effects complementary across actions.

\subsection{Depth-Based Student Policy}
\label{sec:exp:student}

We evaluate the student (Sec.~\ref{sec:method:student}), which must recover the teacher's interaction from egocentric depth and proprioception alone. We ablate the partner-prediction objective that shapes its depth representation, removing its components cumulatively: \emph{(-) future generation} drops the generative prediction of future partner and humanoid motion; \emph{generation $\to$ regression} further drops the generative prediction of past and present partner poses, keeping only the single-step pose-estimation loss $\mathcal{L}_{\text{pose}}$; and \emph{(-) pose prediction} removes partner prediction entirely, training with $\mathcal{L}_{\text{BC}}$ and $\mathcal{L}_{\text{KL}}$.

\begin{table*}[t]
  \centering
  \caption{\textbf{Student policy results.} Contact recall (\%) at 30\,cm on the held-out evaluation split, mean over 5 room layouts.}
  \vspace{-2mm}
  \label{tab:student}
  \begin{tabular}{lccccc}
    \toprule
    Method & Hug & High-five & Hand-shake & Link-arms & Pat-on-back \\
    \midrule
    \rowcolor{LightGreen}
    \textbf{Ours}               & 34.5 & 13.1 & \textbf{82.0} & \textbf{67.8} & \textbf{51.1} \\
    (-) future generation       & \textbf{43.2} & 14.0 & 75.2 & 42.5 & 44.2 \\
    generation $\to$ regression & 31.3 & \textbf{33.9} & 75.2 & 44.5 & 44.4 \\
    (-) pose prediction         & 20.4 & 15.0 & 64.3 & 46.9 & 38.4 \\
    \bottomrule
  \end{tabular}
\vspace{-6mm}
\end{table*}

\begin{wrapfigure}{r}{0.42\linewidth}
\vspace{-5mm}
\centering
\includegraphics[width=\linewidth]{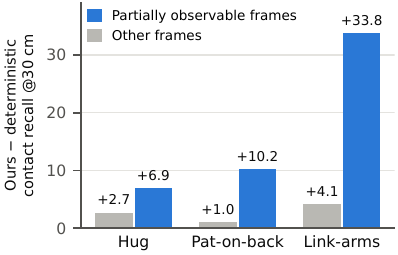}
\vspace{-6mm}
\caption{Within each interaction, frames are split into moments when the partner is hard to see (blue), either too close to fit in the view or largely out of view, and the remaining frames (gray). (Appendix~\ref{sec:supp:occlusion}).}
\label{fig:occlusion}
\vspace{-5mm}
\end{wrapfigure}

\noindent\textbf{Partner-state prediction under occlusion (Q3)}\quad Table~\ref{tab:student} shows that behaviour cloning alone is a weak signal
for the depth encoder: removing partner prediction lowers held-out Contact
Recall on four of five interactions, by $14.1$, $17.7$, $20.9$, and $12.7\,$pt
on hug, hand-shake, link-arms, and pat-on-back. Most of the gain comes from
generating the partner's future, which adds $6.8$ to $25.3\,$pt on hand-shake,
link-arms, and pat-on-back, whereas generating rather than regressing the
current and past poses helps only hug. High-five is the exception, where the
deterministic variant (generation $\to$ regression) is best.
We further systematically investigate where generation helps by relating
contact recall to how much of the view the partner and the robot's own body
occupy (Appendix~\ref{sec:supp:occlusion}). Over the deterministic variant, the
gain concentrates where the partner is least observable
(Fig.~\ref{fig:occlusion}): in close contact on hug and pat-on-back, and on
link-arms, where the partner largely leaves the view, by $33.8\,$pt.

\noindent\textbf{Teacher vs.\ student (Q4)}\quad Compared on the same motions
and thresholds (Appendix~\ref{sec:supp:teacher-student}), the depth student
recovers on average $75\%$ of the privileged teacher's contact recall at
$30\,$cm, and up to $94\%$ on hand-shake. By contrast, replacing depth with an external pose estimator (HMR2.0) collapses contact recall (Appendix~\ref{sec:supp:hpe-compare}).

\begin{figure}[t]
\centering
\setlength{\tabcolsep}{0.6pt}
\renewcommand{\arraystretch}{0.4}
\newcommand{\simqual}[1]{\includegraphics[width=0.19\linewidth]{images/sim-unseen/#1.jpg}}
\begin{tabular}{@{}ccccc@{}}
\small High-five & \small Hand-shake & \small Link-arms & \small Pat-on-back & \small Hug \\[2pt]
\simqual{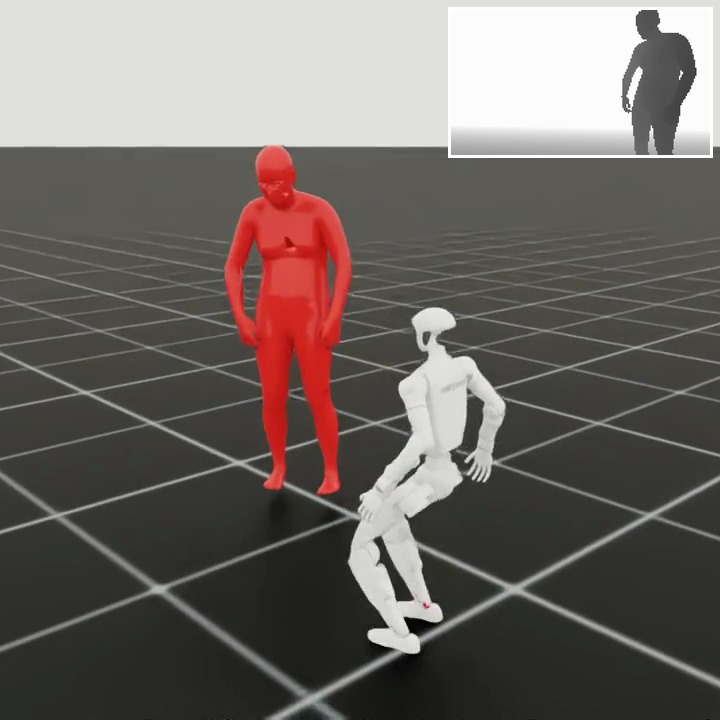} & \simqual{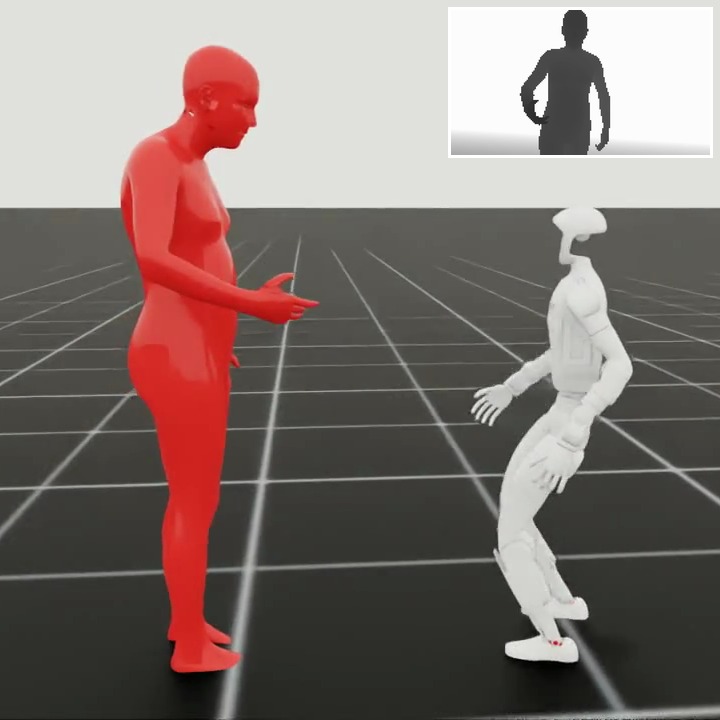} & \simqual{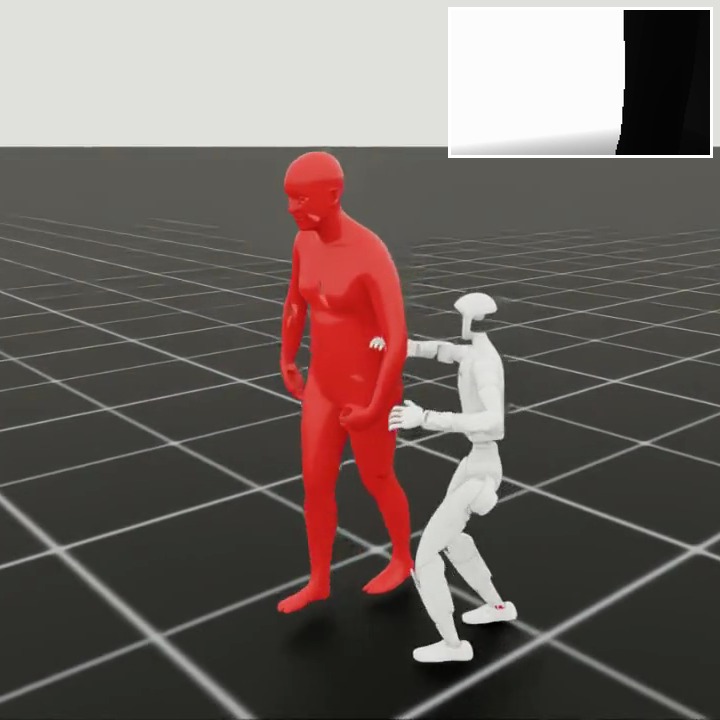} & \simqual{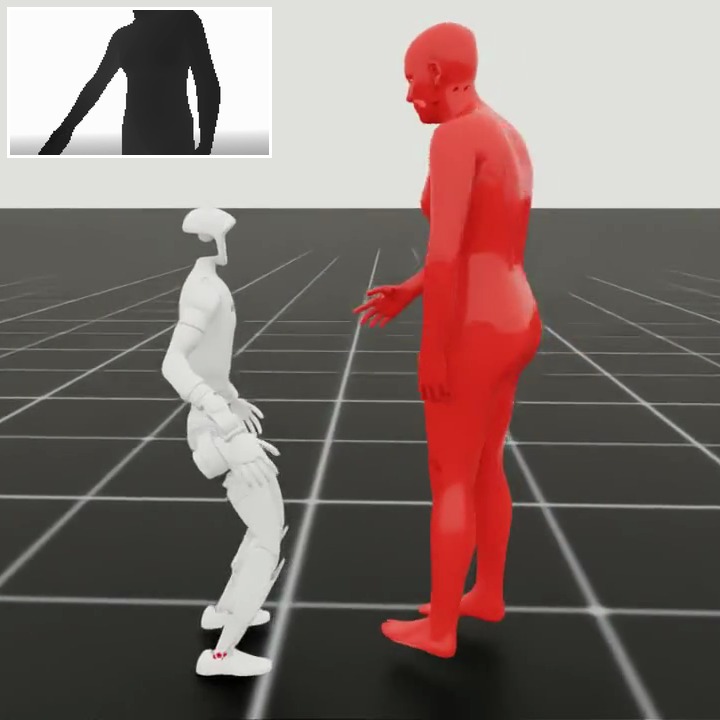} & \simqual{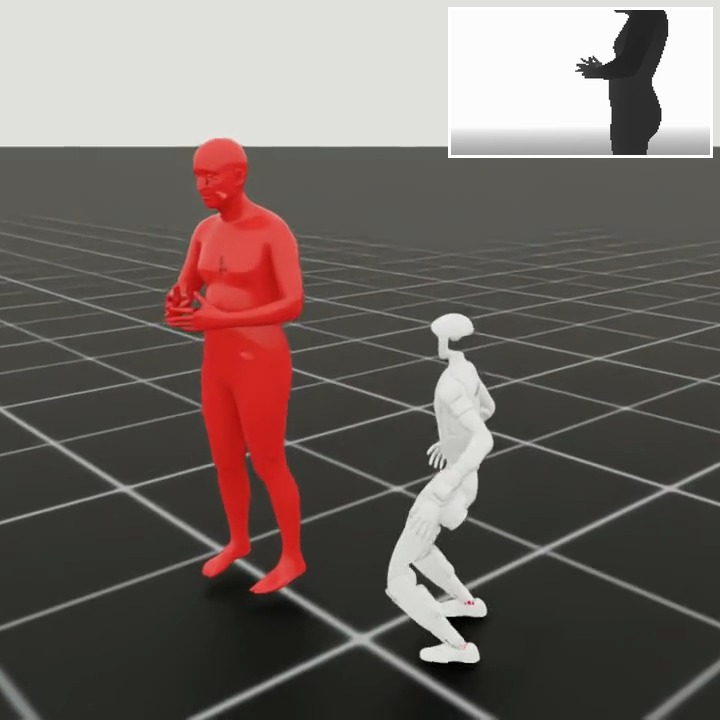} \\
\simqual{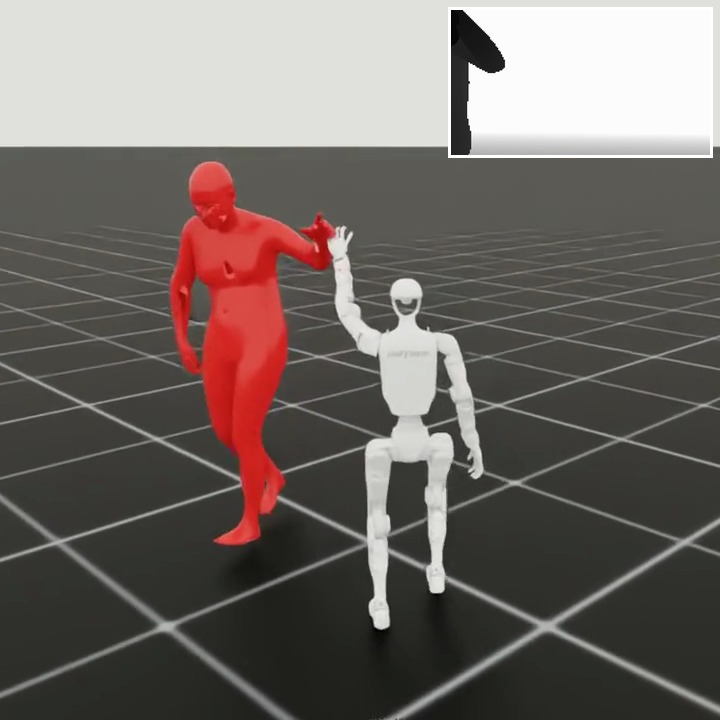} & \simqual{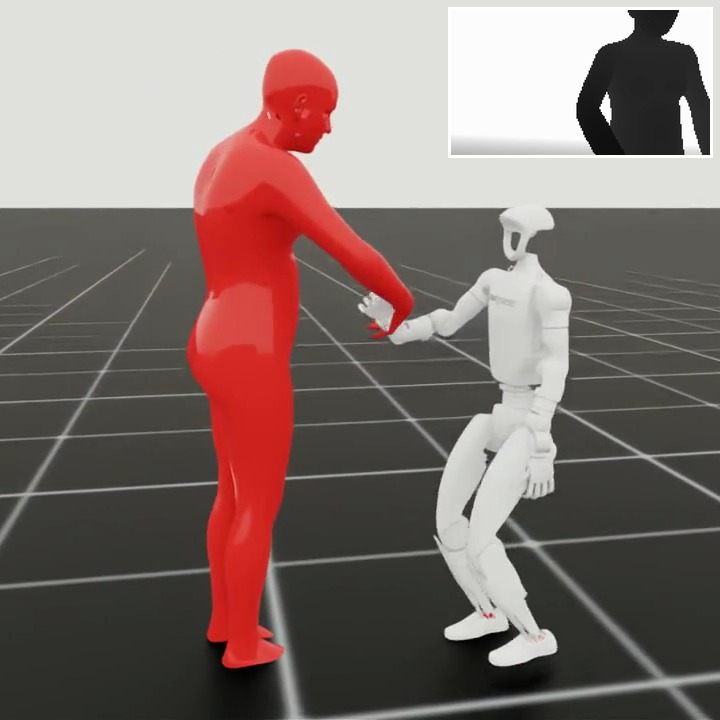} & \simqual{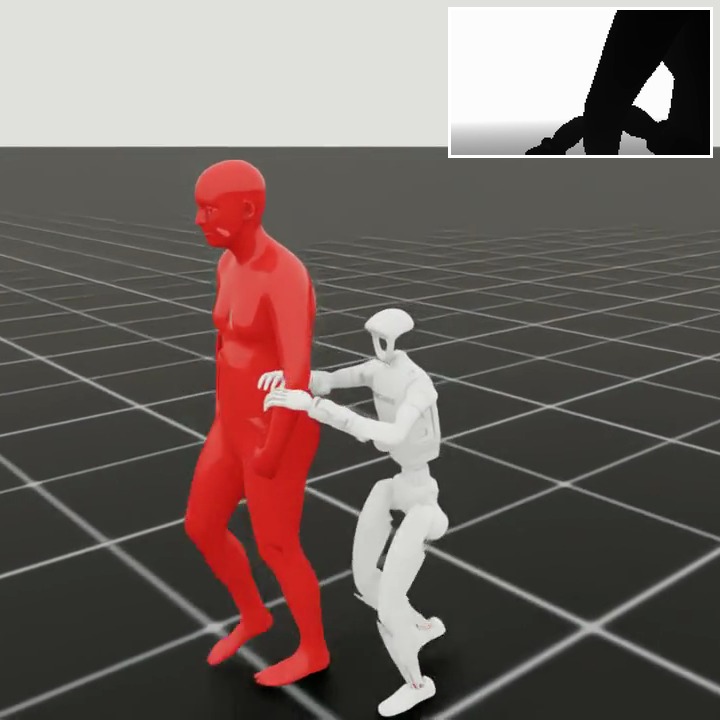} & \simqual{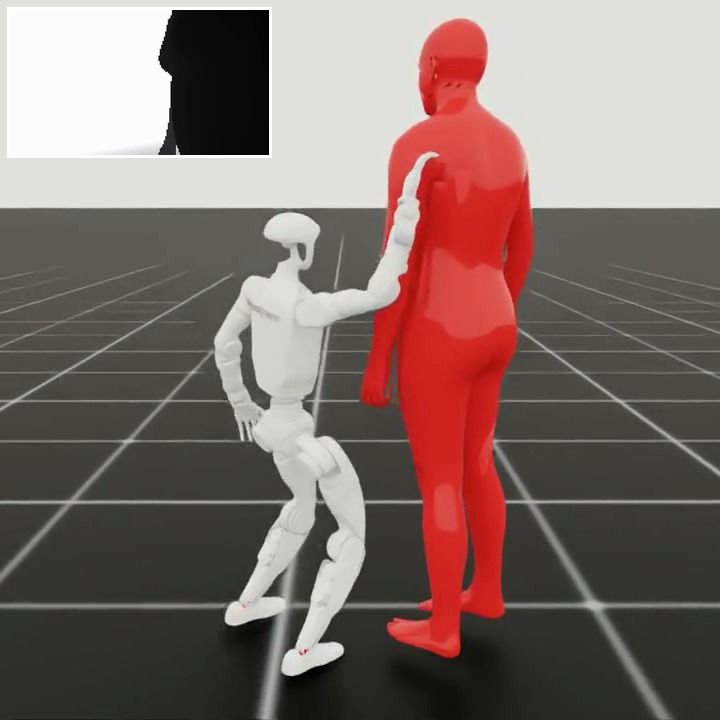} & \simqual{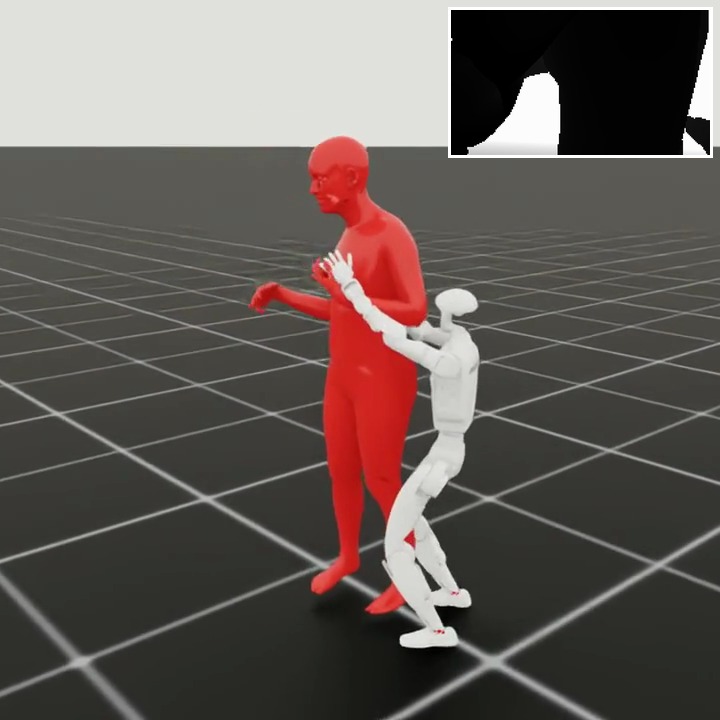} \\
\end{tabular}
\vspace{-2mm}
\caption{\textbf{Student rollouts on held-out motions in simulation} (inset: egocentric depth input)}
\label{fig:sim-unseen}
\vspace{-6mm}
\end{figure}

Figure~\ref{fig:sim-unseen} shows student rollouts on held-out motions never seen during training. Although the partner is frequently truncated or fills the camera view at contact range (insets), the student approaches the partner and establishes the demonstrated contact.

Although we train a separate student per interaction, we treat these students as skills and let a VLM compose them into a single autonomous system: observing only the robot's egocentric RGB and depth, the VLM searches for the partner, selects the skill that matches the partner's configuration, and switches skills when the partner moves (Appendix~\ref{sec:supp:vlm-select}).

\subsection{Real-World Deployment}
\label{sec:exp:realworld}

We deploy the per-interaction students on a physical Unitree~G1 with a human partner across all five interactions. The policies are trained entirely in simulation and run zero-shot, with no real-world fine-tuning; at test time each receives only egocentric depth and proprioception, with no external motion capture or privileged partner state. Because the onboard stereo depth loses the partner's silhouette while the robot moves, we feed the policy metric depth estimated from the onboard RGB image by DepthAnything~3~\cite{lin2025depth} (Fig.~\ref{fig:pipeline}, \emph{Real Deployment}), which keeps the deployed observation close to the rendered training depth (Sec.~\ref{sec:method:rendering}).

\noindent\textbf{Real-robot results (Q5)}\quad Across ten trials per interaction, the students succeed in $31$ of $50$ trials; per-interaction success rates, success criteria, and the model setting used for these trials are in Appendix~\ref{sec:supp:real}. Figure~\ref{fig:teaser} shows the deployment on the real robot, and videos are on the project page. The deployed policies further generalise across partner distances and orientations and sustain repeated interaction, completing three consecutive hugs in one ${\sim}30\,$s episode without reset (Appendix~\ref{sec:supp:real-qual}).



\section{Conclusion}
\label{sec:conclusion}

We presented REACT, a scalable framework for learning deployable humanoid-human interaction policies from large-scale human-human motion-capture data. REACT addresses two key challenges in humanoid-human interaction: the difficulty of learning from noisy human-human demonstrations under heavy mutual occlusion, and the challenge of deploying interaction policies from onboard sensing under severe partial observability. Our framework combines VLM-guided contact-aware retargeting, privileged interaction learning with interaction-aware contact rewards, egocentric depth-policy distillation with photorealistic in-simulator human rendering, and auxiliary generative objectives that predict the partner's pose so that the student can infer the partner's state even under occlusion.

Through extensive simulation experiments and real-world deployment on a Unitree G1 humanoid, we showed that REACT improves preservation of demonstrated interaction contacts and enables robust interaction learning from onboard sensing during close-contact interaction. Our results suggest that scalable humanoid-human interaction may require jointly addressing semantic contact understanding, physically grounded interaction learning, and predictive inference under partial observability. More broadly, we believe REACT provides a step toward deployable humanoid systems capable of safe and adaptive physical interaction with humans in real-world environments. We discuss the limitations of REACT in Appendix~\ref{sec:supp:limitations}.

\clearpage
\acknowledgments{This work was partially supported by JSPS KAKENHI Grant Number 24K22296 and 25H01159, and by JST BOOST, Japan Grant Number JPMJBS2409.}


\bibliography{main,bibnew}

\clearpage
\appendix
\section{Real-World Deployment}
\label{sec:supp:real}
All real-robot results in this section were obtained with an earlier setting of the students, with a three-layer CNN depth encoder instead of ResNet-18.

\subsection{Qualitative results}
\label{sec:supp:real-qual}
\label{sec:real-deploy}

Figure~\ref{fig:teaser} of the main paper shows the real-world deployment of each interaction. Here we provide additional qualitative results.

\begin{figure}[h]
\centering
\includegraphics[width=\linewidth]{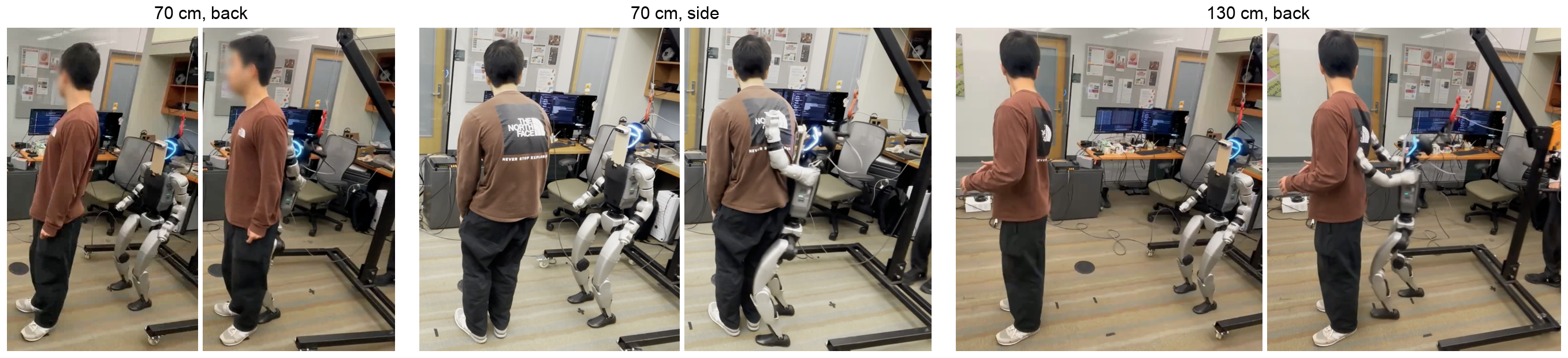}
\vspace{-2mm}
\caption{\textbf{Generalization across partner distance and orientation.} REACT performs pat-on-back with the partner at $70\,\text{cm}$ and $130\,\text{cm}$ and with the partner back-on or side-on to the robot; the per-interaction student reaches and holds the back contact in every condition from onboard depth alone.}
\label{fig:generalization}
\end{figure}

The deployed policy is also robust to where the partner stands and how they are oriented. The pat-on-back student establishes the back contact across partner distances ($70$ and $130\,\text{cm}$) and orientations (back-on and side-on to the robot), reaching and holding the contact in every condition (Figure~\ref{fig:generalization}).

\begin{figure}[t]
\centering
\includegraphics[width=\linewidth]{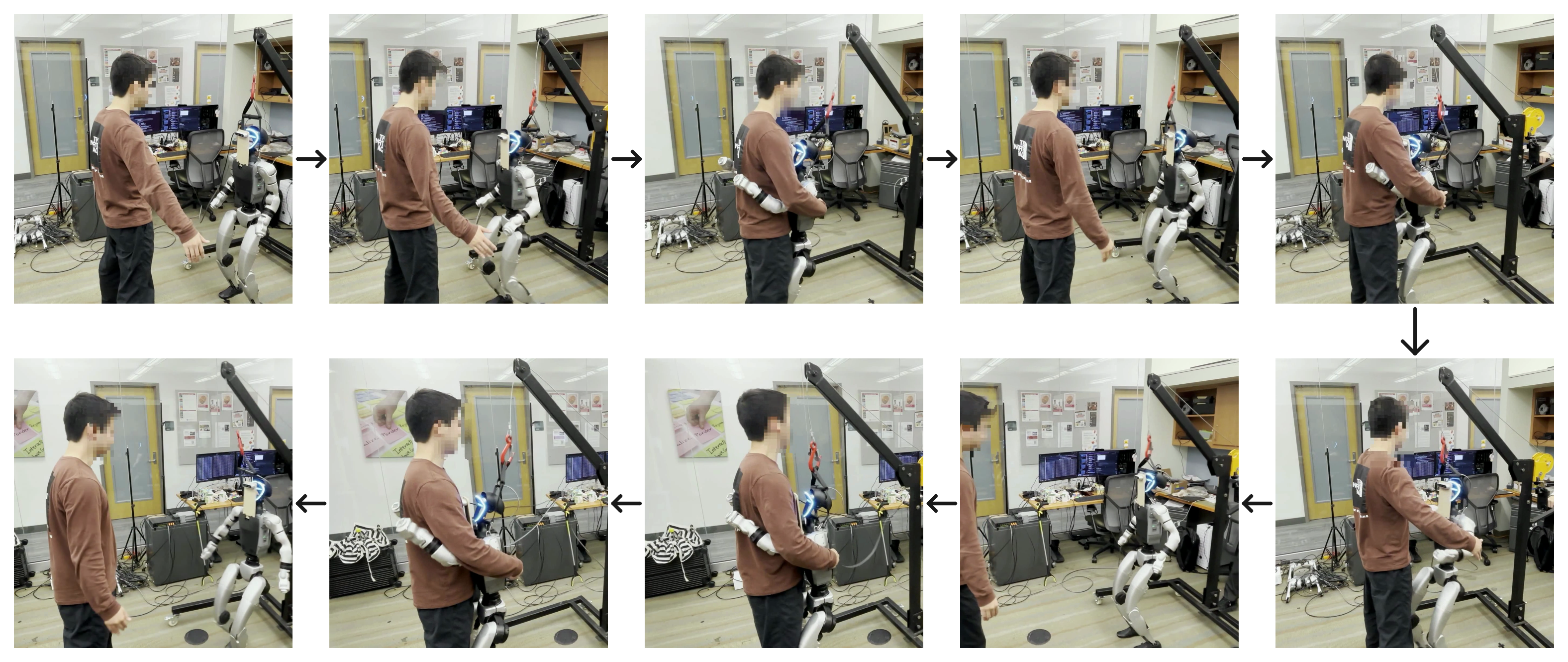}
\vspace{-4mm}
\caption{\textbf{Long-horizon behavior: three consecutive hugs in a single episode.} A continuous ${\sim}30\,\text{s}$ rollout in which the policy completes the hug interaction three times in a row, with no reset or re-initialisation between repetitions. The ten snapshots are read in boustrophedon order, following the arrows. After each embrace the robot releases the partner, re-approaches, and re-establishes contact, so the three hug cycles appear as three separate close-contact phases across the panels. The policy sustains repeated contact-rich interaction over a long horizon rather than a single one-shot contact.}
\label{fig:longhorizon-hug}
\vspace{-4mm}
\end{figure}

Beyond a single contact event, the same policy sustains the interaction over a long horizon: it performs the hug three times in a row within one continuous rollout, with no reset, no re-initialisation, and no operator intervention between repetitions (Figure~\ref{fig:longhorizon-hug}). Each cycle runs the full interaction loop, in which the robot approaches the partner, establishes and briefly holds the embrace, then releases, steps back to a stable stance, and re-approaches to initiate the next hug.
Completing three such cycles consecutively within a ${\sim}30\,\text{s}$ episode demonstrates that the behavior is not a one-off contact event but rather a robust and repeatable closed-loop skill. The policy reliably recovers a stable post-contact state and re-engages from it, enabling repeated interaction cycles without degradation after the first successful execution.

\subsection{Quantitative results}
\label{sec:supp:real-quant}
We evaluate each per-interaction student over ten real-world trials (Table~\ref{tab:realworld}). The success criterion depends on the interaction: for hand-shake and high-five, where the partner actively raises and lowers a hand, a trial succeeds if the robot reacts to this motion and brings its hand within roughly $15\,\text{cm}$ of the partner's; for hug, pat-on-back, and link-arms, the partner stands still at varied positions, and a trial succeeds if the robot approaches the partner and completes the interaction. Deployed zero-shot, the students succeed in $31$ of $50$ trials overall, with hug and pat-on-back the most reliable at $8/10$ each.

\begin{table}[h]
  \centering
  \small
  \caption{\textbf{Real-world success rate.} Successful / total trials per interaction on the physical Unitree~G1, ten trials per interaction.}
  \label{tab:realworld}
  \vspace{2mm}
  \begin{tabular}{cccccc}
    \toprule
    High-five & Hug & Hand-shake & Link-arms & Pat-on-back & Total \\
    \midrule
    5/10 & 8/10 & 5/10 & 5/10 & 8/10 & 31/50 \\
    \bottomrule
  \end{tabular}
\end{table}

\subsection{Deployment-time depth processing}
\label{sec:supp:deploy-depth}
The student is trained on the in-simulator SMPL-X mesh depth described in Sec.~\ref{sec:supp:rendering}, which provides clean human geometry and sharp
body silhouettes. On the physical robot, the onboard Intel RealSense D435 does
not recover these silhouettes reliably: while the robot is moving, its stereo
depth becomes noisy and loses the partner's contour exactly where the policy
needs it. We therefore do not use the RealSense depth stream at deployment. Instead we run monocular
metric depth estimation with DepthAnything~3 (Large, metric)~\cite{lin2025depth} on the onboard RGB
image and feed the estimated depth to the policy. Inference runs in real time
on the onboard compute through a TensorRT/DeepStream
pipeline.
This keeps the deployed observation close to the sharp, contour-rich depth the
student was trained on, while requiring only a single RGB camera. The demonstration videos on the project page instead use Fast-FoundationStereo~\cite{wen2026fastfoundationstereo} on the RealSense left and right infrared images.

\subsection{Safe Interaction through Force Regulation}
\label{sec:supp:safe-interaction}
In close-contact interaction the magnitude of the force the robot applies to the person is a safety concern. We make this force observable and controllable during training by spawning the human partner as a physically simulated, collidable body that tracks the reference motion through PD control rather than a purely kinematic reference, so that every contact the policy makes produces a real reaction force on the corresponding G1 link. The force-regulation reward $r^{\text{force}}_t$ of Sec.~\ref{sec:supp:contact-reward} (Eq.~\ref{eq:contact-force}) then shapes how hard the robot touches. It rewards contact forces inside a comfortable band and penalises force at links that should not be in contact. Training the teacher with this term drives the policy toward gentle, bounded contact instead of forceful pushing, so the resulting interaction stays soft enough to be comfortable for the human while remaining firm enough to be felt. The deployable student inherits this behavior through distillation, imitating the actions of the force-regulated teacher.

\section{Composing Per-Interaction Students with VLM Skill Selection}
\label{sec:supp:vlm-select}

We train one student per interaction. This is a design choice rather than a limit on generalisation: adding a new interaction costs one additional policy instead of retraining all of them. To run the students as a single autonomous system, we treat them as a skill library and add a VLM-based selection layer on top, evaluated in simulation. The VLM agent observes only the robot's own egocentric RGB and depth, never the partner's state, and decides both where the robot should move and which student should take control.

\begin{figure}[t]
\centering
\includegraphics[width=0.75\linewidth]{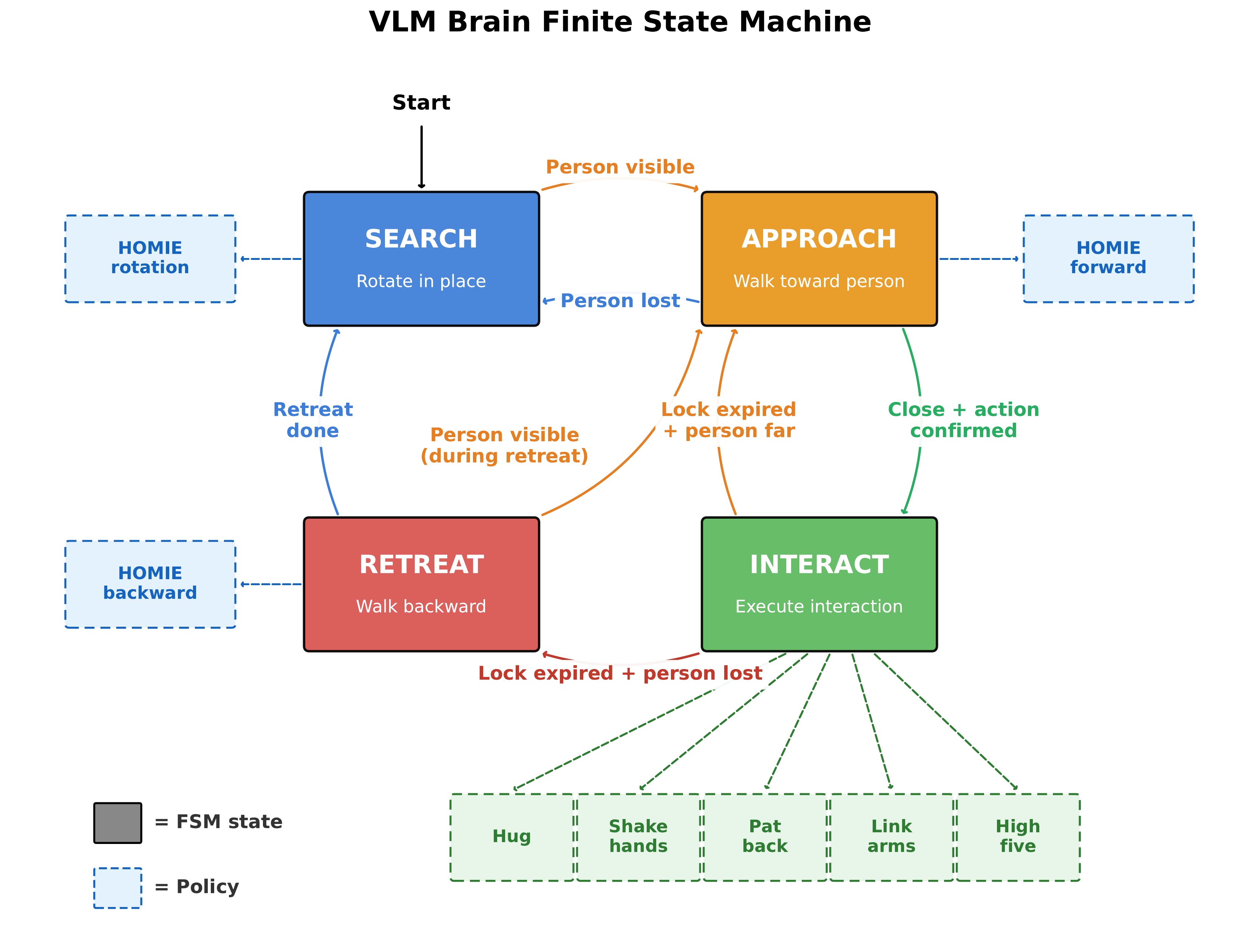}
\vspace{-2mm}
\caption{\textbf{VLM skill-selection state machine.} Solid boxes are states and dashed boxes are the policies they invoke: the HOMIE locomotion controller for Search, Approach, and Retreat, and one of the five per-interaction students for Interact.}
\label{fig:supp:vlm-fsm}
\end{figure}

\paragraph{Finite state machine} The selection layer is a four-state machine (Figure~\ref{fig:supp:vlm-fsm}). In \emph{Search}, the robot rotates in place through the HOMIE locomotion controller until a person becomes visible. In \emph{Approach}, it walks toward the person and returns to Search if the person is lost. Once the robot is close and the VLM has confirmed which interaction the partner's configuration calls for, the machine enters \emph{Interact} and hands control to the matching student (hug, hand-shake, pat-on-back, link-arms, or high-five). The selected interaction is locked for a fixed period; when the lock expires, the robot re-approaches if the partner is still visible but far, or enters \emph{Retreat} and walks backward if the partner is lost, after which it resumes Search or re-approaches as soon as the partner reappears.

\begin{figure}[t]
\centering
\includegraphics[width=\linewidth]{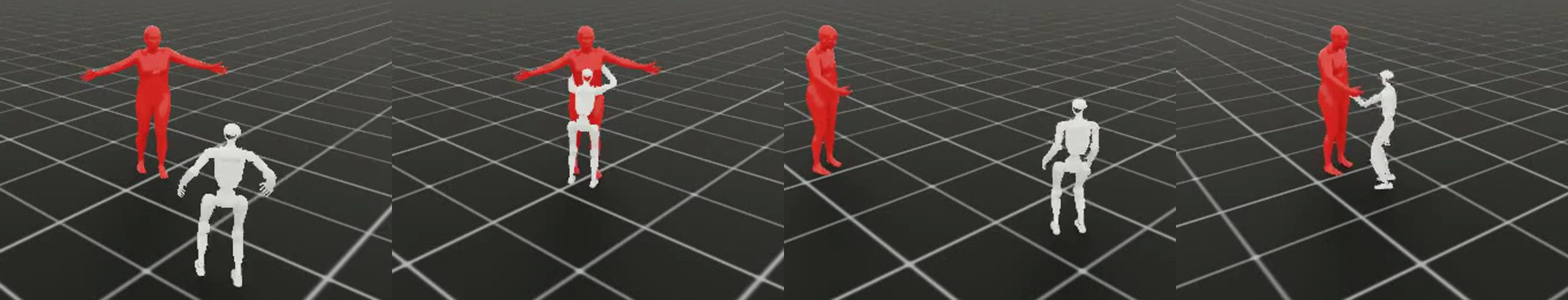}
\vspace{-5mm}
\caption{\textbf{One continuous rollout with VLM skill selection.} No interaction is specified in advance. From left to right: the partner opens their arms, and the VLM selects the hug student; the partner is then displaced mid-rollout, and the robot re-finds them, approaches, and switches to the hand-shake student, all without a reset.}
\label{fig:supp:vlm-strip}
\end{figure}

\paragraph{Qualitative result} Figure~\ref{fig:supp:vlm-strip} shows one continuous rollout in which no interaction is specified. The robot first selects and executes a hug from the partner's open-arm posture. After the partner is displaced mid-rollout, the robot re-finds the partner, approaches, and selects the hand-shake student, without any reset or operator intervention. A safe abort for a partner who pulls away mid-contact is left to future work.

\section{Student Policy}

\subsection{Realistic Human Depth Rendering}
\label{sec:supp:rendering}
\begin{figure}[t]
\centering
\includegraphics[width=\linewidth]{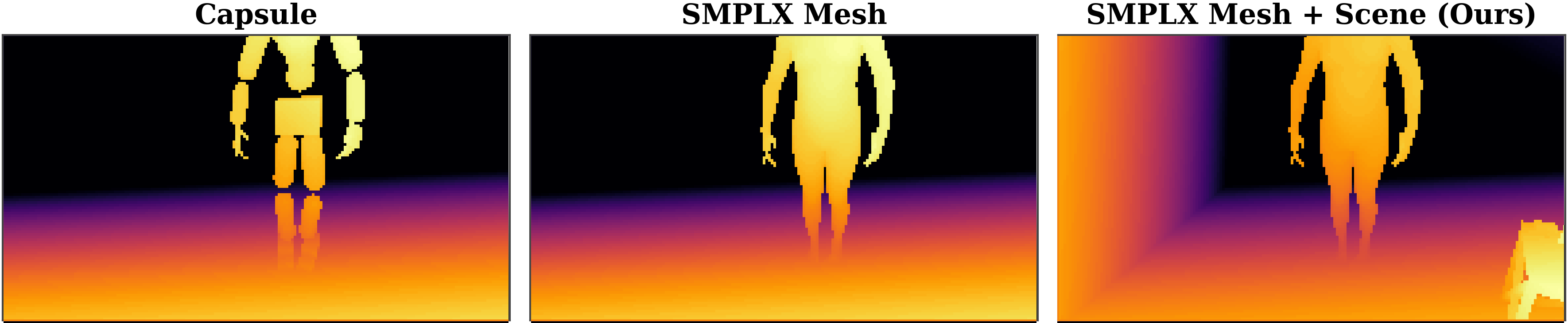}
\vspace{-4mm}
\caption{\textbf{In-loop human depth rendering closes the sim-to-real gap.}
Egocentric depth of a hug frame. \emph{Capsule}: the
partner as collision primitives, blocky and non-human. \emph{SMPL-X Mesh}: the
partner rendered as a realistic body mesh. \emph{SMPL-X Mesh + Scene (Ours)}: our
full rendering with the randomised scene. This last variant closely resembles the
real-sensor depth shown in Figure~1 of the main paper.}
\label{fig:depth-compare}
\vspace{-4mm}
\end{figure}

A depth policy transfers to the real robot only if its training depth resembles what the onboard sensor will see, so closing the sim-to-real depth gap is essential. Physics simulators represent the human partner with capsule colliders, whose egocentric depth (Fig.~\ref{fig:depth-compare}, left) looks nothing like a real person. We hide these capsules from the depth camera and, for rendering only, attach a visual SMPL-X body mesh that we update in the loop, so that at every control step a single batched GPU SMPL-X forward pass advances every environment's mesh to its current frame and the camera sees a realistic moving human (Fig.~\ref{fig:depth-compare}, middle). The capsule colliders stay in the physics scene so contact is still simulated, while the mesh is used only for rendering. To the best of our knowledge, this is the first system to model physical interaction with a human in a physics simulator that simultaneously renders the human as a photorealistic SMPL-X mesh for perception and simulates contact through capsule colliders. The surrounding scene is further randomised with mesh assets refreshed during training, so the depth also carries realistic background geometry (Fig.~\ref{fig:depth-compare}, right). This in-loop human-and-scene depth is what lets the student deploy zero-shot on the real robot.

\subsection{Why End-to-End Depth Beats Pose-Estimation-Based Methods}
\label{sec:supp:hpe-compare}

A natural way to drive interaction with a human partner is to attach an external human-pose-estimation module, recover the partner's body from the camera stream, and feed the estimated pose to the policy as the partner-state input~\cite{huang2026beyond,goel2023humans4d}. This design factors perception from control and reuses mature off-the-shelf pose estimators, but it breaks down in the close-range regime that physical interaction requires.
At contact range, the partner is largely occluded by the robot's own body or leaves the camera field of view, so an external estimator either loses the partner or returns a stale pose precisely when accurate partner state matters most.

REACT instead relies on no external pose-estimation module: it maps the egocentric depth image to action end-to-end, and uses partner pose estimation only as an auxiliary task that improves training efficiency.

Table~\ref{tab:supp-hpe-compare} shows the quantitative comparison between these two approaches. We build the external-pose baseline following 4D~Humans~\cite{goel2023humans4d}. At each step, a person detector localizes the partner and HMR2.0 reconstructs the partner's body. The recovered partner joint positions are concatenated with the robot's proprioception and fed to the policy, replacing the egocentric depth input that REACT uses. This substitution collapses mesh-contact recall on every interaction, falling from $78.6$ to $8.2$ on hand-shake and from $15.3$ to $5.4$ on high-five, because the external estimator degrades under exactly the self-occlusion and field-of-view loss that close, contact-rich interaction induces. The end-to-end student, by contrast, stays reliable in this regime, where an external estimator is least able to recover the partner state.

\begin{figure}[t]
\centering
\includegraphics[width=\linewidth]{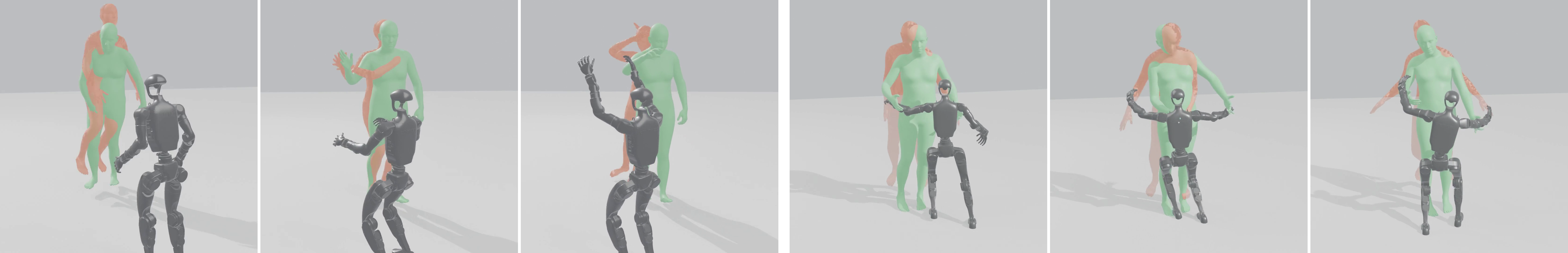}
\vspace{-4mm}
\caption{\textbf{Why the external pose estimation-based method fails at contact range.} Temporally ordered snapshots of an external-pose rollout, read left to right, with the ground-truth partner (\textcolor{green}{green}, \emph{Human (Sim)}) and the HMR2.0 reconstruction (\textcolor{red}{red}, \emph{HMR2.0 Prediction}) overlaid against the robot (black). \textbf{Left three:} high-five; as the partner's hand leaves the robot's egocentric field of view, the reconstructed hand position deviates sharply from the ground truth at the moment of contact. \textbf{Right three:} hug; at the instant of embrace the partner's torso occludes the camera, and the reconstruction collapses into a large error behind the true mesh. The estimator degrades precisely in the close-contact, self-occluded regime that REACT's end-to-end student handles directly from depth.}
\label{fig:hpe-failure}
\vspace{-4mm}
\end{figure}

To better understand why the external-pose baseline fails, Figure~\ref{fig:hpe-failure} overlays the ground-truth partner mesh (green) and the HMR2.0 reconstruction (red) in the same scene as the robot, for a high-five and a hug episode.
The two interactions expose complementary failure modes of external pose estimation at contact range.
In the high-five case (left), the partner raises a hand toward the robot, and as that hand crosses the edge of the robot's egocentric field of view, the reconstructed arm jumps to a completely wrong configuration. Such an implausible pose drives the policy's partner-state input out of the distribution it was trained on, so the policy reaches to the wrong location. In the hug case (right), the partner remains in view during the approach, but at the instant of embrace, the partner's torso fills and occludes the camera. With almost no visible body surface to fit, HMR2.0 produces a large, unstable error and the reconstruction breaks apart behind the true mesh.
In both cases the estimator degrades exactly when the interaction enters close contact, which is when accurate partner state matters most, and this is the mechanism behind the recall collapse in Table~\ref{tab:supp-hpe-compare}.

\begin{table}[t]
  \centering
  \small
  \caption{\textbf{End-to-end vs.\ external human-pose estimation.}
  Mesh-contact recall at a $30\,\text{cm}$ threshold, in percent, on the held-out
  evaluation split.
  \emph{Ext.\ human pose (HMR2.0)} drives the policy from partner state produced
  by an external human-pose estimator while \emph{Ours (E2E)} outputs actions end-to-end from egocentric depth and proprioception. Both rows use earlier model settings than Table~\ref{tab:student} and are evaluated on the same held-out motions.}
  \label{tab:supp-hpe-compare}
  \vspace{2mm}
  \begin{tabular}{lccccc}
    \toprule
    Method & Hug & Hand-shake & Pat-on-back & Link-arms & High-five \\
    \midrule
    Ext.\ human pose (HMR2.0~\cite{goel2023humans4d}) & 5.7 & 8.2 & 1.3 & 3.9 & 5.4 \\
    \rowcolor{LightGreen}
    Ours (E2E)              & \textbf{38.0} & \textbf{78.6} & \textbf{23.8} & \textbf{30.1} & \textbf{15.3} \\
    \bottomrule
  \end{tabular}
  \vspace{-2mm}
\end{table}

\subsection{Depth-based Student Policy: architecture, losses, and distillation}
\label{sec:supp:student}

\paragraph{Observations} At every control step the student receives (i) a 5-frame egocentric depth stack at offsets $\{0, 0.5, 1.0, 1.5, 2.0\}\,\text{s}$ into the past, (ii) a 5-step short-term proprioception history ($5 \times 108 = 540$ dims), and (iii) a $100$-step ($2\,\text{s}$ at $50\,\text{Hz}$) long-term proprioception history. 

\paragraph{Depth backbone} A shared ResNet~\cite{He_2016_CVPR} embeds each of the five depth frames independently. The per-frame embeddings are concatenated into a temporal depth representation.

\paragraph{Long-term proprio encoder} A 1D-CNN maps the $100$-step proprioception window to a compact embedding. 

\paragraph{Conditional VAE} Let $\mathbf{c}_t$ concatenate the temporal depth embedding, the short-term proprioception history, and the long-term proprioception embedding. A \textit{prior} MLP maps $\mathbf{c}_t$ to a diagonal Gaussian $\mathcal{N}(\boldsymbol{\mu}_p, \boldsymbol{\sigma}_p^2)$; a training-only \textit{encoder} MLP maps the current reference joint positions and velocities $\mathbf{o}^{\text{ref}}_t$ to $\mathcal{N}(\boldsymbol{\mu}_e, \boldsymbol{\sigma}_e^2)$. We combine the two heads residually, $\boldsymbol{\mu}_z = \boldsymbol{\mu}_p + \boldsymbol{\mu}_e$, and a decoder MLP maps $[\mathbf{c}_t, \mathbf{z}_t]$ to the same 18-dim HOMIE-plus-arm action mean as the teacher. At deployment the encoder is dropped and $\mathbf{z}_t$ is drawn from the prior $\mathcal{N}(\boldsymbol{\mu}_p, \boldsymbol{\sigma}_p^2)$.

\paragraph{Pose-anticipation auxiliary head} A DDPM denoiser conditioned on the temporal depth embedding and the long-term proprioception embedding predicts the partner's past, current, and short-horizon future SMPL-X key-body state (positions, velocities, and orientations), $P^{\text{partner}}_t$, together with the G1's own near-future trajectory, $P^{\text{humanoid}}_t$, over 5 past/current frames at $\{0, 0.5, 1.0, 1.5, 2.0\}\,\text{s}$ and 3 future frames at $\{0.5, 1.0, 1.5\}\,\text{s}$ ahead. The objective sums a multi-step DDIM forward-prediction loss and a GENMO-style~\cite{li2025genmo} single-step pose-estimation loss in equal proportion, and the combined pose loss enters the total distillation loss with weight $0.5$, i.e.\ $\lambda_{\text{gen}} = \lambda_{\text{pose}} = 0.5$. Per-frame losses are gated by how much of the human partner is visible in the depth image. Frames in which the partner occupies $15\%$ or less of the depth pixels contribute no gradient.

\paragraph{Distillation loss} The full distillation loss defined in the main paper combines an L2 behavior-cloning term against the teacher action, the conditional-VAE KL between the privileged encoder posterior and the deployable prior (with $\lambda_{\text{KL}}$ ramped from $10^{-4}$ to $10^{-2}$ between epochs 500 and 2000), and the two pose objectives above. We use on-policy DAgger~\cite{ross2011dagger}.

\paragraph{Other student training details} The egocentric camera is a simulated Intel RealSense D435 mounted on the G1's upper body and held level, so it faces forward rather than down at the floor and keeps the partner inside the frustum during close interaction. The stock G1 mount is pitched about $48^{\circ}$ downward for locomotion, so on the real robot we use a custom level mount that the simulated camera matches. It runs at $30\,\text{Hz}$ with an $87^{\circ}$ horizontal field of view and a $3\,\text{m}$ far clip, renders at $192{\times}108$ without down-sampling, and these intrinsics match the physical D435 we deploy. Because such an $87^{\circ}$-HFOV camera cannot see a partner who is, in the demonstration, beside or behind the humanoid, we restrict the training and evaluation corpus to clips in which the partner stays within the frustum. We train on 128 parallel Isaac Sim environments on a single L40S GPU with the partner driven by a reference PD controller and rendered as in Sec.~\ref{sec:supp:rendering}. Demonstration-time augmentation includes human motion speed warping $[0.8, 1.2]\times$, a random start-time offset up to $1\,\text{s}$, a random G1 spatial offset of $0.15\,\text{m}$, a random yaw offset up to $0.32\,\text{rad}$ ($\approx 18.5^\circ$), and segment slowdown up to $2\times$ over windows of up to $3\,\text{s}$. Depth-image augmentation applies edge displacement, distance-dependent additive Gaussian and sparkle noise, background-occlusion rectangles, and random rectangular and corner masks, and the deepest pixels are clamped to the $3\,\text{m}$ far clip. We distil a separate student per interaction, so each of the five actions has its own deployable checkpoint.

\subsection{Gain of generative prediction under partial observability}
\label{sec:supp:occlusion}

Fig.~\ref{fig:occlusion} of the main paper compares the full student (\emph{Ours}) with the deterministic variant (\emph{generation $\to$ regression} in Table~\ref{tab:student}) as a function of how much of the egocentric view is occupied. Both are evaluated on the held-out split with the indoor scene randomisation (walls, ceiling, and furniture) disabled, so that every depth pixel above the floor and nearer than the $3\,$m far clip belongs to the partner or to the robot's own body. For every contact-labelled frame we record this occupancy, the fraction of such pixels in the depth image, and whether the contact is satisfied at $30\,$cm.

The partially observable regime differs by interaction. On hug and pat-on-back it is close contact, where the two bodies fill the view, and we split at an occupancy of $0.8$. On link-arms it is the partner leaving the view, and we split at $0.4$, since fewer than $2\%$ of its contact frames exceed $0.8$. Both thresholds were set after inspecting the occupancy distributions. Table~\ref{tab:supp-occlusion} reports the recall difference in each regime over three evaluation seeds, which change the simulation seed while keeping the same checkpoints.

\begin{table}[h]
  \centering
  \small
  \caption{\textbf{Contact recall gain of Ours over the deterministic variant (pt at 30\,cm) by occupancy.} Scene randomisation disabled, held-out split. PO: partially observable regime. Share: fraction of contact frames in the regime.}
  \label{tab:supp-occlusion}
  \begin{tabular}{llccc}
    \toprule
    Interaction & Regime & Share & Seeds 1 / 2 / 3 & Mean \\
    \midrule
    \multirow{2}{*}{Hug}         & PO (occupancy $\geq 0.8$) & 26\% & $+7.4$ / $+5.7$ / $+7.7$   & $\mathbf{+6.9}$ \\
                                 & otherwise                 & 74\% & $+2.3$ / $+4.2$ / $+1.6$   & $+2.7$ \\
    \midrule
    \multirow{2}{*}{Pat-on-back} & PO (occupancy $\geq 0.8$) & 22\% & $+12.4$ / $+12.1$ / $+6.2$ & $\mathbf{+10.2}$ \\
                                 & otherwise                 & 78\% & $+0.1$ / $-1.6$ / $+4.6$   & $+1.0$ \\
    \midrule
    \multirow{2}{*}{Link-arms}   & PO (occupancy $< 0.4$)    & 70\% & $+35.9$ / $+34.1$ / $+31.4$ & $\mathbf{+33.8}$ \\
                                 & otherwise                 & 30\% & $+10.9$ / $-6.1$ / $+7.6$  & $+4.1$ \\
    \bottomrule
  \end{tabular}
\end{table}

In every seed and on all three interactions, the gain is larger in the partially observable regime. The seed-paired difference between the two regimes is significant on link-arms ($t(2)=5.6$, $p<0.05$) and consistent in sign but not significant on hug ($t(2)=3.1$) and pat-on-back ($t(2)=2.4$). Three caveats apply. The seeds vary only the evaluation, with one trained checkpoint per variant. Occupancy counts the robot's own body, which also occludes the partner. And occupancy results from each policy's own motion, so the analysis is correlational rather than an intervention on visibility.

\subsection{Teacher vs.\ student on matched motions and thresholds}
\label{sec:supp:teacher-student}

The main paper scores the teacher at 15\,cm on all motions and the student at 30\,cm on held-out motions. To compare them directly, Table~\ref{tab:supp:teacher-student} evaluates the teacher (the distillation source) and the deployed student on the same motions, the student's training split, since the teacher is trained on all motions and has no held-out split. Both use the same evaluation protocol and per-frame mesh contact recall. The teacher values differ slightly from Table~\ref{tab:teacher} because they are computed on the student's training motions and from a single seed. Averaged over the five interactions, the student recovers $52\%$ and $75\%$ of the teacher's recall at 15 and 30\,cm, respectively.

\begin{table}[h]
  \centering\small
  \caption{\textbf{Teacher vs.\ student contact recall (\%) on matched motions.} T: privileged teacher, S: egocentric depth student, Rec.: S/T. The mean recovery averages the per-interaction ratios.}
  \label{tab:supp:teacher-student}
  \begin{tabular}{lcccccc}
    \toprule
    & \multicolumn{3}{c}{15\,cm} & \multicolumn{3}{c}{30\,cm} \\
    \cmidrule(lr){2-4}\cmidrule(lr){5-7}
    Interaction & T & S & Rec. & T & S & Rec. \\
    \midrule
    Hug         & 52.3 & 18.1 & 35\% & 80.2 & 49.9 & 62\% \\
    High-five   & 69.5 & 23.9 & 34\% & 87.7 & 62.1 & 71\% \\
    Hand-shake  & 93.9 & 78.6 & 84\% & 96.7 & 91.1 & 94\% \\
    Link-arms   & 72.5 & 44.1 & 61\% & 89.9 & 74.1 & 82\% \\
    Pat-on-back & 87.0 & 38.4 & 44\% & 93.9 & 61.4 & 65\% \\
    \midrule
    Mean        & 75.0 & 40.6 & 52\% & 89.7 & 67.7 & 75\% \\
    \bottomrule
  \end{tabular}
\end{table}

\section{Contact Label Filtering and Surface Projection}

\subsection{VLM Contact Annotation: distance candidates and semantic verification}
\label{sec:supp:contact-annotation}

We annotate five contact-rich Inter-X~\cite{interx} actions: hug, hand-shake, pat-on-back, high-five, and link-arms by following these steps shown below.

\paragraph{Stage 1: distance candidates} We run SMPL-X forward kinematics on both bodies and, for every other frame, group each person's arm mesh vertices by skinning weights into a \emph{hand} (wrist and fingers), \emph{forearm}, and \emph{upper-arm} region on each side. Each region's vertices are compared against every vertex of the partner mesh; pairs within $15\,\text{cm}$ are grouped by (contacting arm region, contacted SMPL-X joint), the closest pair is retained per group, and detections separated by gaps of up to $0.5\,\text{s}$ are merged into events. Events shorter than $0.1\,\text{s}$ are discarded. Each event records its time window, the contacting region, the contacted partner joint, the closest actor- and partner-side vertices, and the minimum and average mesh-to-mesh distances.

\begin{figure}[t]
\centering
\includegraphics[width=0.92\linewidth]{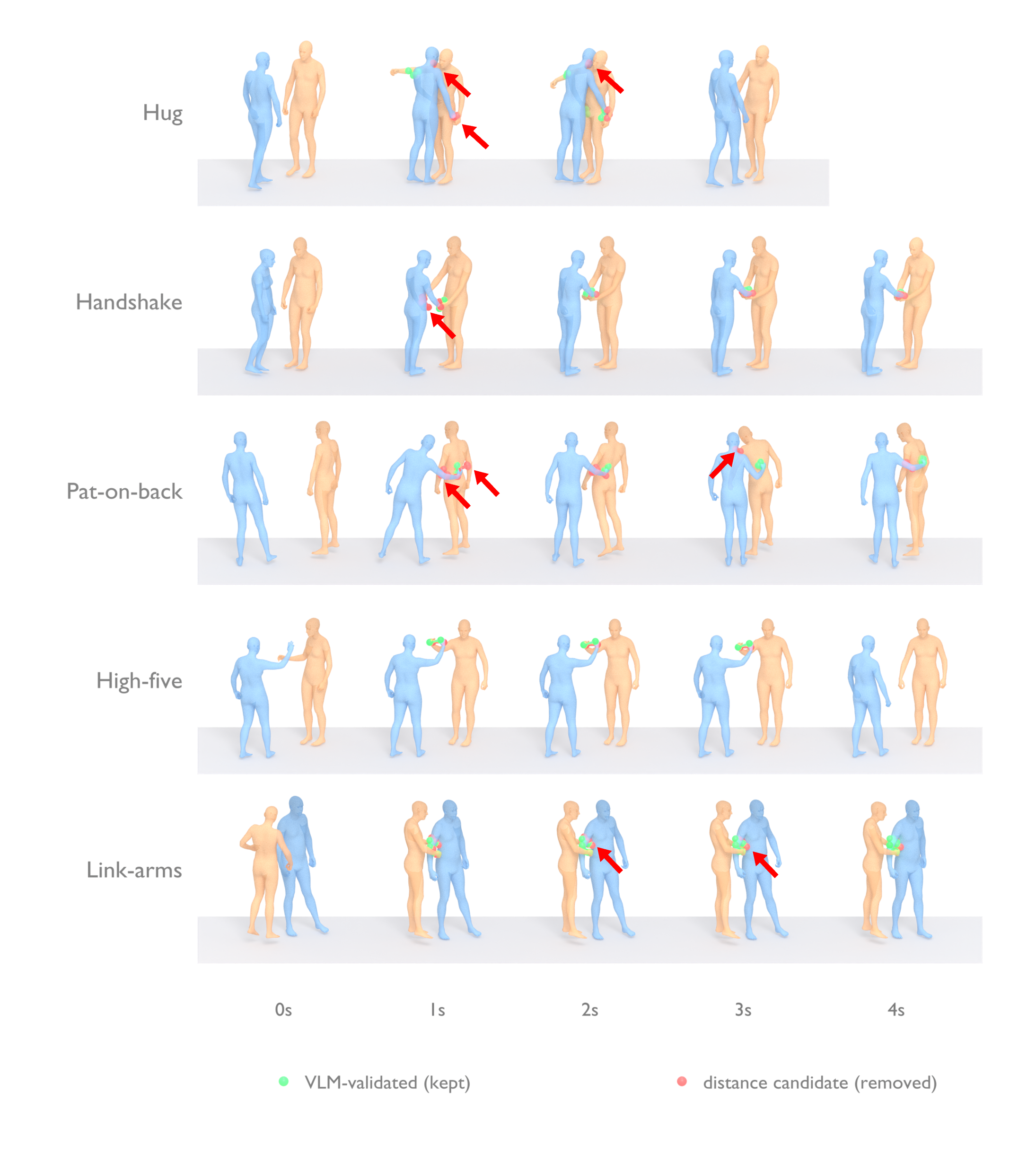}
\caption{\textbf{VLM contact filtering, qualitative examples.} Distance candidates and their VLM verdicts on the five contact-rich Inter-X actions (rows: hug, hand-shake, pat-on-back, high-five, link-arms), sampled at $1\,\text{s}$ intervals over a $4\,\text{s}$ window (columns). Green markers are candidates kept after VLM verification, labelled essential or incidental. Red markers are Stage~1 distance candidates the VLM removes as spurious near-contacts. The proximity test fires on many transient arm passes; the VLM stage prunes them while preserving the genuine contacts, which is the precision gain quantified in Tab.~\ref{tab:supp:vlm-label-eval}.}
\label{fig:supp:vlm-contact}
\end{figure}

\paragraph{Stage 2: VLM verification} Each event is verified against the corresponding video by a vision-language model (Gemini~2.0 Flash). We render the SMPL-X motion from three camera viewpoints with person-1 in blue and person-2 in orange, concatenate the views horizontally with per-view foreground cropping, sample at $5\,\text{fps}$, and split the video into $3\,\text{s}$ chunks; each chunk is sent with the action label and a textual description of every candidate contact event that overlaps it. The VLM assigns each contact candidate one of four semantic labels (\emph{essential}, \emph{incidental}, \emph{unnecessary}, or \emph{avoid}), together with the contacting and contacted surfaces (palm/inner vs.\ back/outer) and the views in which the contact is clearly visible. Only events labelled essential or incidental survive; Figure~\ref{fig:supp:vlm-contact} shows representative kept and removed candidates across the five actions. The verbatim prompt is given in Sec.~\ref{sec:supp:vlm-prompt}.

\paragraph{Output} For each of the six contact sites (left or right \emph{hand}, \emph{forearm}, \emph{upper-arm}) and every timestep $t$, the annotation provides a binary contact flag, the closest vertex on the actor mesh, the contacted target vertex on the partner mesh, and the name of the contacted SMPL-X joint. The contact reward of Eq.~\ref{eq:contact-reward} and the Stage~2 contact co-optimisation in retargeting consume these flags for all six sites, mapped to the G1 wrist, elbow, and shoulder links.

\paragraph{Label-quality evaluation against manual ground truth}
To isolate the contribution of Stage~2, we evaluate the contact \emph{labels} themselves, independently of any retargeting, against a hand-annotated ground-truth set. We manually annotate $80$ Inter-X clips ($16$ per action, five actions: hug, hand-shake, pat-on-back, high-five, link-arms), marking the actor SMPL-X vertex, the contacted partner SMPL-X vertex, and the time window of each true contact event. We then treat the Stage~1 distance candidates and the Stage~2 VLM-verified events as predicted events and match them to GT events on a per-frame basis. At every frame, an actor--partner contact pair is counted as a true positive iff (i) the actor vertices of the GT and the prediction lie within $\epsilon = 10\,\text{cm}$ on the SMPL-X rest-pose template (matching the retargeting threshold used in the main paper), and (ii) the same holds on the partner side. To avoid double-counting the same physical contact when the predictor independently emits its mirrored (partner-actor) candidate, we deduplicate predictions whose vertex pair coincides with that of an existing same-frame actor--partner candidate. We report per-clip precision, recall, and $F_1$ as the across-clip mean $\pm$ standard deviation, both per action and on the full $80$-clip set.

\begin{table*}[t]
  \centering
  \small
  \caption{\textbf{VLM contact-label quality on $80$ manually-annotated Inter-X clips ($16$ per action).} Per-action and total precision, recall, and $F_1$ as across-clip mean $\pm$ std under per-frame vertex-distance matching at $\epsilon = 10\,\text{cm}$ on the SMPL-X rest-pose template, after symmetric-pair deduplication. \emph{Ours w/o VLM filtering} uses the raw Stage~1 distance candidates; \emph{Ours} additionally applies VLM verification. Reporting precision and recall separately shows that the VLM gain is almost entirely a \emph{precision} gain, while recall is essentially preserved (or only mildly reduced). The verifier removes spurious near-contacts rather than acting as a stricter detector.}
  \label{tab:supp:vlm-label-eval}
  \setlength{\tabcolsep}{4pt}
  \begin{tabular}{l ccc ccc}
    \toprule
    & \multicolumn{3}{c}{Ours w/o VLM filtering (distance)} & \multicolumn{3}{c}{Ours (VLM-verified)} \\
    \cmidrule(lr){2-4} \cmidrule(lr){5-7}
    Action & P\,$\uparrow$ & R\,$\uparrow$ & $F_1$\,$\uparrow$ & P\,$\uparrow$ & R\,$\uparrow$ & $F_1$\,$\uparrow$ \\
    \midrule
    Hug         & $0.060_{\pm 0.05}$ & $0.241_{\pm 0.22}$ & $0.127_{\pm 0.08}$ & $0.081_{\pm 0.10}$ & $0.201_{\pm 0.21}$ & $\mathbf{0.181}_{\pm 0.13}$ \\
    Hand-shake   & $0.275_{\pm 0.15}$ & $0.785_{\pm 0.38}$ & $0.495_{\pm 0.10}$ & $0.442_{\pm 0.33}$ & $0.661_{\pm 0.41}$ & $\mathbf{0.674}_{\pm 0.21}$ \\
    Pat-on-back & $0.107_{\pm 0.07}$ & $0.824_{\pm 0.34}$ & $0.212_{\pm 0.09}$ & $0.227_{\pm 0.12}$ & $0.760_{\pm 0.35}$ & $\mathbf{0.387}_{\pm 0.12}$ \\
    High-five   & $0.260_{\pm 0.14}$ & $0.866_{\pm 0.32}$ & $0.417_{\pm 0.16}$ & $0.336_{\pm 0.27}$ & $0.660_{\pm 0.46}$ & $\mathbf{0.626}_{\pm 0.17}$ \\
    Link-arms   & $0.257_{\pm 0.16}$ & $0.600_{\pm 0.31}$ & $0.374_{\pm 0.18}$ & $0.339_{\pm 0.20}$ & $0.549_{\pm 0.34}$ & $\mathbf{0.453}_{\pm 0.18}$ \\
    \midrule
    \rowcolor{LightGreen}
    Total ($80$ clips) & $0.192_{\pm 0.15}$ & $\mathbf{0.663}_{\pm 0.39}$ & $0.330_{\pm 0.19}$ & $\mathbf{0.285}_{\pm 0.25}$ & $0.567_{\pm 0.41}$ & $\mathbf{0.468}_{\pm 0.23}$ \\
    \bottomrule
  \end{tabular}
\end{table*}

Splitting $F_1$ into precision and recall (Table~\ref{tab:supp:vlm-label-eval}) makes the role of the VLM stage explicit and shows that the actions divide into two distinct regimes. On the total set the VLM stage raises across-clip mean precision by $48\%$ ($0.192{\to}0.285$) while recall only drops mildly ($0.663{\to}0.567$), so the $+42\%$ $F_1$ gain ($0.330{\to}0.468$) is precision-driven. The verifier mainly removes spurious candidates from the Stage~1 set rather than acting as a stricter detector.

The first regime covers pat-on-back, high-five, hand-shake, and link-arms, where the distance candidates already recall most true contacts, with recall ranging from $0.60$ to $0.87$, but at very low precision of only $0.11$ to $0.28$, because close-range arm motion fires many spurious near-contacts, which is where the low distance-only $F_1$ comes from. VLM verification raises precision by $1.3$--$2.1\times$ (e.g.\ pat-on-back $0.107{\to}0.227$, hand-shake $0.275{\to}0.442$) while recall drops by $0.05$--$0.21$, lifting $F_1$ the most. Hug is the opposite case. Its recall is already low for both methods, $0.24$ and $0.20$, so the low $F_1$ is recall-bound rather than precision-bound. The cause is structural, since $33\%$ of hug GT events have the actor on the torso/pelvis, outside the six hand/forearm/upper-arm sampling regions that both stages enumerate, so the predictor cannot emit those candidates at all. VLM verification cannot recover contacts that were never proposed, so its hug gain is small and comes from cleaning the few arm-region candidates rather than from added coverage.

\paragraph{Annotation tool and ground truth} The $80$ GT clips are annotated with a custom GUI that records, for each true contact event, the actor SMPL-X vertex, the contacted partner SMPL-X vertex, and the start/end frames. GT events are authored in an actor-centric convention and remapped to the Inter-X canonical P1/P2 frame before scoring (the same remap used in Sec.~\ref{sec:supp:retarget-eval}); without this remap roughly half of the clips would compare the wrong person. Each clip is annotated by a single expert.

\subsection{Human-to-Robot Surface Correspondence Registration}
\label{sec:supp:registration}

\begin{figure}[t]
\centering
\includegraphics[width=\linewidth]{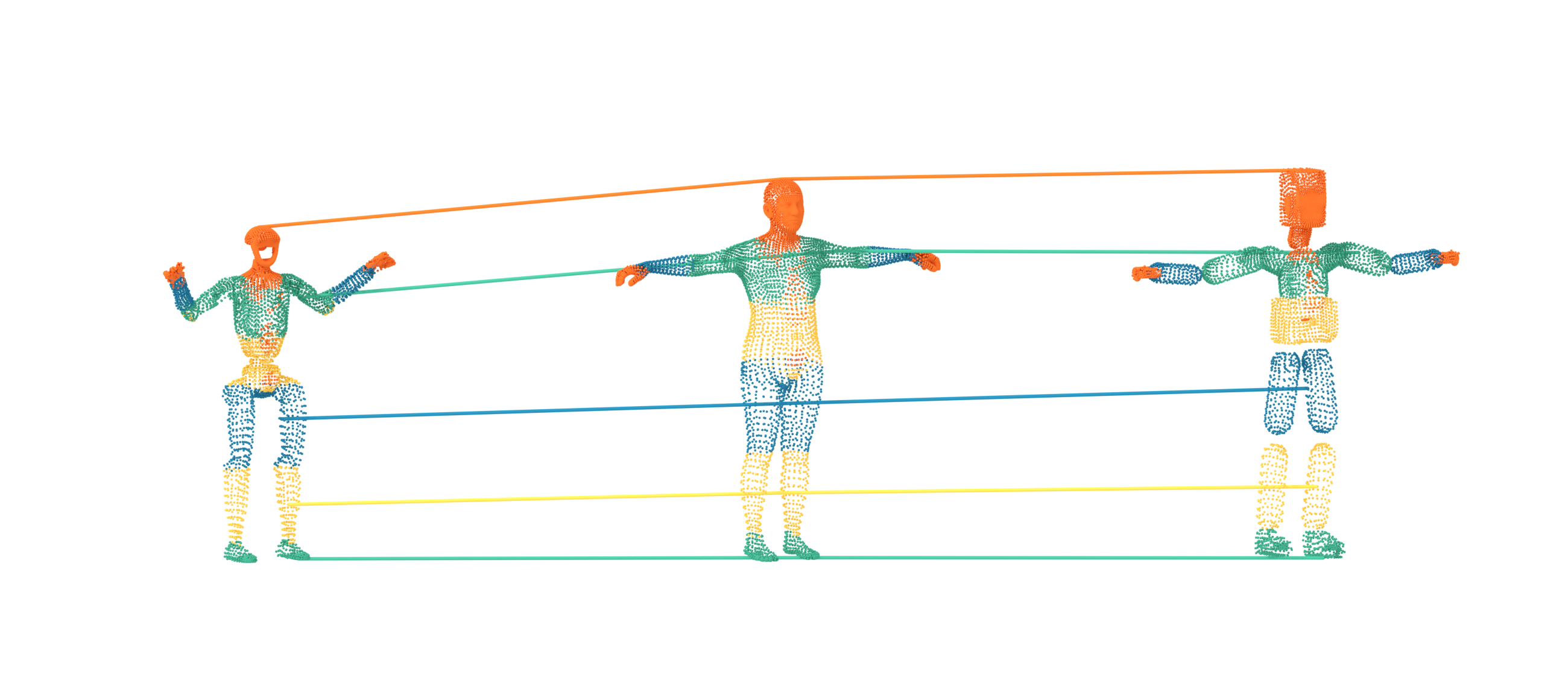}
\caption{\textbf{Body correspondence projects the SMPL-X mesh onto either the G1 or the partner-capsule surface.} The offline registration introduced in Sec.~\ref{sec:supp:registration} establishes a dense body correspondence that maps every SMPL-X body vertex to a $(\text{link},\,\mathbf{p}_\text{local})$ pair, so the same mesh can be projected onto either target surface, the Unitree~G1 robot or the partner's SMPL-X collision capsules. The three columns show one identical set of colour-coded SMPL-X vertices in three configurations, from left to right: projected onto the G1 robot surface, in their SMPL-X template positions, and projected onto the partner capsule surface. Each colour denotes the G1 link a vertex is registered to, and the connecting lines join the same representative vertices across the three configurations, so that any contact vertex can be carried onto the corresponding robot or capsule surface by forward kinematics. The columns are drawn at true relative scale, exposing the proportion mismatch between the human template and the shorter G1 that the registration must bridge. The dense map covers the full template at a mean vertex-to-surface residual of $0.89\,\text{mm}$, well below the $\epsilon = 10\,\text{cm}$ contact threshold.}
\label{fig:supp:smplx2g1}
\end{figure}

We construct an offline map from every SMPL-X body vertex to a $(\text{link},\, \mathbf{p}_\text{local})$ pair on the Unitree~G1 zero-pose surface. At run time, the world-space location of any SMPL-X contact vertex used in Stage~2 retargeting (Sec.~\ref{sec:supp:retarget-opt}) and in the contact reward 
is recovered by forward kinematics on G1 from the stored link-local point. The registration runs in four stages on the SMPL-X mean-shape body and the G1 zero-pose surface mesh.

\paragraph{Stage 1: limb-direction initialisation} A scale-invariant initial alignment between the SMPL-X body and the G1 zero pose is obtained by matching cosines between corresponding limb directions on the two bodies. Using direction rather than length yields a robust initialisation despite the substantial proportion mismatch between the human body and the humanoid robot.

\paragraph{Stage 2: weighted Procrustes alignment} A weighted Procrustes solve over corresponding skeletal joints further aligns the two articulations while preserving their global geometric consistency, providing a clean reference frame for the subsequent surface-level deformation.

\paragraph{Stage 3: Laplacian-regularised non-rigid ICP} We deform the SMPL-X mesh onto the G1 surface with a Laplacian-regularised non-rigid ICP, establishing dense surface-level correspondences. Two safeguards prevent erroneous matches across unrelated body parts despite the surface-proximity criterion. A \emph{decreasing-stiffness schedule} establishes coarse rigid-body-part structure before allowing fine-grained deformations, and \emph{ray-cast normal projection} restricts each correspondence query to surfaces visible along the outward SMPL-X vertex normal, blocking matches that would otherwise wrap around the robot from the opposite side.

\paragraph{Stage 4: post-snap and storage} A final post-snap stage projects the remaining outlier vertices, those whose Stage~3 target falls in free space or on a kinematically unrelated link, onto the nearest valid robot surface point, ensuring complete surface coverage. For each SMPL-X vertex we store the resulting G1 link identifier together with the link-local position $\mathbf{p}_\text{local}$.

\paragraph{Runtime use} At every control step, the contact reward of Eq.~\ref{eq:contact-reward} and retargeting (Sec.~\ref{sec:supp:retarget-opt}) retrieve the G1-side contact point as $\mathbf{p}^h_t = T_{\text{link}^h}(\boldsymbol{q}_t)\,\mathbf{p}_{\text{local}^h}$, where $T_{\text{link}^h}(\boldsymbol{q}_t)$ is the forward-kinematics transform of the stored link at joint configuration $\boldsymbol{q}_t$ and $(\text{link}^h, \mathbf{p}_{\text{local}^h})$ is the registered pair for contact site $h$.

\paragraph{Registration quality} Across the full SMPL-X template the recovered correspondence attains a mean vertex-to-surface residual of $0.89\,\text{mm}$ on the G1 zero-pose mesh, far below the $\epsilon = 10\,\text{cm}$ contact threshold used throughout evaluation (Sec.~\ref{sec:supp:retarget-eval}); registration error is therefore negligible relative to the metric tolerance. Figure~\ref{fig:supp:smplx2g1} visualises the colour-coded link assignment recovered by this procedure.

\section{Contact-Aware Retargeting}
\label{sec:supp:retarget-pipeline}

This section covers the two-stage contact-aware retargeting that closes the demonstrated contacts on the humanoid, given the surface registration and VLM contact labels described in Sec.~\ref{sec:supp:registration} and Sec.~\ref{sec:supp:contact-annotation}. We first show a qualitative comparison of the retargeted motions in Sec.~\ref{sec:supp:retarget-qual}, then detail the two-stage retargeting optimisation itself in Sec.~\ref{sec:supp:retarget-opt}. We describe the dataset and its filtering in Sec.~\ref{sec:supp:data}, and report event-level contact recall and precision against the manual annotations in Sec.~\ref{sec:supp:retarget-eval} together with a human preference study in Sec.~\ref{sec:supp:userstudy}.

\subsection{Qualitative results}
\label{sec:supp:retarget-qual}
\begin{figure}[t]
\centering
\includegraphics[width=\linewidth]{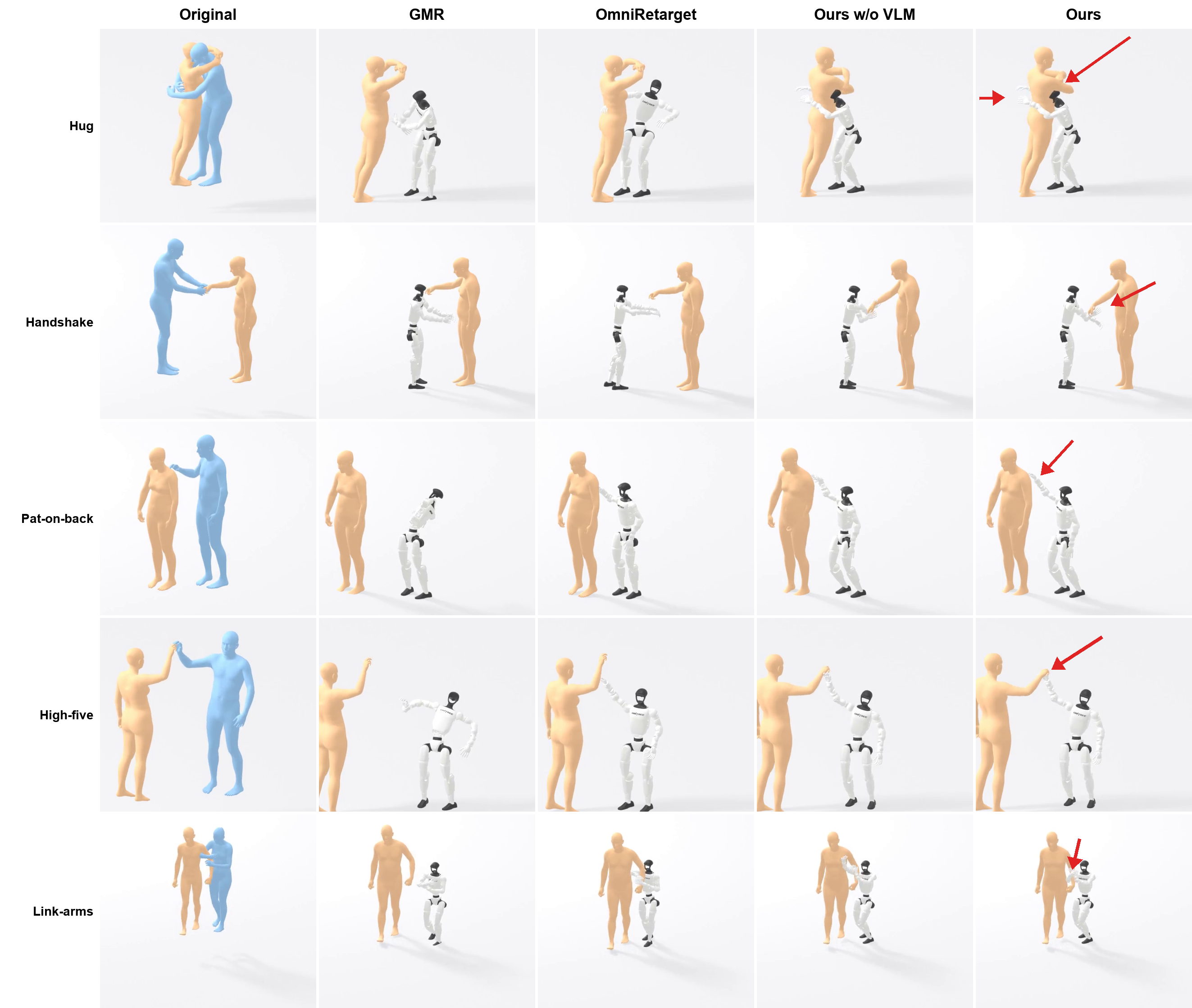}
\caption{\textbf{Qualitative retargeting comparison.} Rows: the five contact-rich actions (hug, hand-shake, pat-on-back, high-five, link-arms). Columns: the original human--human demonstration and four retargets (GMR, OmniRetarget, Ours w/o VLM filtering, Ours), each at the action's peak-contact frame. The red arrow marks the demonstrated contact the retarget must reproduce on the G1. Ours establishes the touch on the robot surface where the baselines reach short or push apart.}
\label{fig:supp:retarget-qual}
\end{figure}

We begin the evaluation with a qualitative comparison of the retargeted motions (Figure~\ref{fig:supp:retarget-qual}). For each of the five contact-rich actions we show the original human--human demonstration alongside the retargeting baselines and our method at the action's peak-contact frame, with an arrow marking the demonstrated contact the retarget must reproduce on the G1. \emph{Ours} establishes the demonstrated touch on the robot surface, whereas the baselines either fail to reach the contact or push the bodies apart to satisfy their solvers, and \emph{Ours w/o VLM filtering} additionally chases spurious distance candidates.

\subsection{Contact-Aware Motion Retargeting}
\label{sec:supp:retarget-opt}
We retarget each human--human demonstration onto the humanoid in two stages. \emph{Stage~1 (kinematic retargeting)} height-scales only the humanoid-side body along gravity and, on contact-labelled frames, blends the arm end-effector targets toward the demonstrated contact positions, so the touches are anchored rather than averaged away. It builds on OmniRetarget~\cite{yang2025omniretarget}. \emph{Stage~2 (contact co-optimisation)}, detailed next, closes the residual contact gap.

\paragraph{Co-optimisation} Stage~2 refines the Stage-1 kinematic initialisation by
\emph{alternating inverse kinematics} between the humanoid and the partner, rather than by
descending a single weighted-sum loss. We run $3$ outer co-optimisation iterations. Each
iteration first solves a humanoid IK step and then a partner SMPL-X IK step, holding the
Stage-1 locomotion (root trajectory and leg motion) fixed throughout.

\paragraph{Targets} Each annotated interaction is converted into per-frame IK targets of two
kinds. The first kind is a \emph{binary contact}, a touch that the VLM filter has verified. For
a binary contact the humanoid contact site $h$ is driven toward a meeting point
$\mathbf{m}^h_t = (1-w)\,cg_h(\mathbf{q}_t) + w\,rc_j$, where $cg_h$ is the G1 link-surface
point, $rc_j$ the targeted partner SMPL-X vertex, and $w$ a slack weight, with
$\mathbf{m}^h_t$ projected onto the partner surface mesh. The second kind is a \emph{proximity
pair}, a body part that stays near the partner without touching and must keep its demonstrated
separation. We flag a proximity pair whenever a partner arm or hand vertex approaches a non-arm
G1 link within $0.08\,\text{m}$, and we drive the humanoid site toward the \emph{distance-preserving}
point $rc_j + \hat{\mathbf{d}}\,d^{\text{ref}}$, where $\hat{\mathbf{d}}$ is the current
partner-to-G1 direction and $d^{\text{ref}}$ the demonstrated distance. This distance-preserving
proximity field keeps the interaction configuration from collapsing. Binary targets enter the
humanoid IK with weight $1.0$, scaled by a $2$-frame approach ramp, and proximity targets with
weight $0.3$.

\paragraph{Solvers} The humanoid IK is a damped least-squares step over the arm DOFs ($10$ inner iterations, orientation weight $0.3$, $N_{\text{sub}}=5$
sub-steps). The partner SMPL-X IK is a gradient step (learning rate $0.003$, $100$ inner
iterations, pose-regularisation weight $0.003$ toward the Stage-1 pose, temporal-smoothness
weight $0.02$). Interpenetration is suppressed by a \emph{volumetric SMPL signed-distance
field}~\cite{ICCV25:VolumetricSMPL} evaluated against the G1 collision mesh (collision weight
$10.0$, margin $0.005\,\text{m}$), making the IK collision-aware. At most $8$ contact pairs per frame
are optimised, and clips are truncated to $10\,\text{s}$.

\paragraph{Contact-adaptive temporal smoothing} We smooth the optimised arm body-pose and
arm joint DOFs with a Gaussian temporal filter whose width adapts to how much contact the clip
contains. The effective width is $\sigma_{\mathrm{eff}} = \sigma\,\min(1,\, n_c/n_0)$ with base
width $\sigma = 2$ frames, $n_c$ the number of contact frames, and $n_0$ a contact-count
threshold, so dense-contact clips are smoothed at the full width while sparse-contact actions
such as high-five receive proportionally less smoothing and keep their brief contact peaks. The
contact source is the VLM-verified set for \emph{Ours} and the raw distance candidates for
\emph{Ours w/o VLM filtering}. All other settings are identical between the two.

\subsection{Dataset Statistics and Filtering}
\label{sec:supp:data}
We draw from five contact-rich Inter-X action categories: hug, hand-shake, pat-on-back,
high-five, and link-arms. The master pool of reconstructed human--human sequences contains
$1355$ clips (hug $282$, hand-shake $278$, pat-on-back $275$, high-five $252$, link-arms $268$).

\paragraph{Teacher pool (1100)} We curate the master pool using the Inter-X action label
attached to each clip by our VLM annotation: clips whose realised motion does not match the
nominal interaction (e.g.\ lift-up and other off-label variants that are infeasible or
unsafe to reproduce on the G1) are removed. This drops $255$ clips ($18.8\%$) and leaves
$N{=}1100$ valid sequences (hug $244$, hand-shake $233$, pat-on-back $196$, high-five $205$,
link-arms $222$). This $1100$-clip pool is the teacher training corpus and also the pool on
which Jitter and Penetration are reported (Sec.~\ref{sec:supp:retarget-eval}); the
$80$ manually GT-annotated clips used for Recall/Precision are a subset of it.

\paragraph{Student pool (human-scored subset)} For the deployable student we apply a second,
stricter curation: two human raters score the retargeted result of each candidate clip on a
$3$-point scale, and only clips that both rate at the top score ($3$) are kept, so the student
trains on the cleanest contact reproductions. The student additionally restricts to the
FOV-filtered subset (Sec.~\ref{sec:supp:student}), in which the partner stays inside the
torso-camera frustum for the entire horizon, and uses a $9{:}1$ train/evaluation split.

\subsection{Retargeting metric definitions}
\label{sec:supp:retarget-eval}

We detail the four metrics used in the main paper's retargeting evaluation, namely contact Recall, Precision, Jitter, and Penetration.
The ground truth is the set of $80$ manually annotated clips of
Sec.~\ref{sec:supp:contact-annotation}: each GT contact event $e$
specifies an actor SMPL-X mesh vertex $v_e^{\text{act}}$, a partner
SMPL-X mesh vertex $v_e^{\text{par}}$, and a time window
$[t_e^{\text{s}}, t_e^{\text{e}}]$. The contact threshold is
$\epsilon = 10\,\text{cm}$ throughout; results at other thresholds
are reported in Table~\ref{tab:supp:retarget-sweep}. Both metrics are averaged
across-clip with $\pm$ standard deviation.

\paragraph{Recall (vertex-pair-targeted, event-binary)}
For each GT event $e$ we evaluate the distance between the two
annotated mesh vertices: the actor vertex $v_e^{\text{act}}$ is
mapped to a point on the G1 surface through the SMPL-X--G1
registration introduced in  Sec.~\ref{sec:supp:registration}; the partner vertex
$v_e^{\text{par}}$ is taken from the partner's SMPL-X forward
kinematics at frame $t$. At each frame $t$ in the annotated window we
measure the distance between the two mapped vertices, where $g$ denotes
the registration-followed-by-G1-FK mapping and $h$ the partner SMPL-X FK,
and take the closest approach over the window,
\begin{equation}
  d_e = \min_{t\in[t_e^{\text{s}}, t_e^{\text{e}}]}
  \big\| g(v_e^{\text{act}}, t) - h(v_e^{\text{par}}, t) \big\|.
\end{equation}
The event is counted as recovered when this closest distance falls below
the threshold, that is when $d_e < \epsilon$. Per-clip Recall is the
fraction of recovered GT events, and we report the across-clip mean.

\paragraph{Precision (body-part-centric, event-matching)}
We partition the SMPL-X mesh into $11$ coarse body parts
(\emph{head, torso, pelvis, left/right \{arm, hand, leg, foot\}})
by SMPL-X skinning-weight argmax; the same labels apply to the
partner mesh and, through the registration, to the humanoid surface.
For each ordered pair of body parts $(b_{\text{act}},
b_{\text{par}})$ and each frame $t$ we compute the per-vertex
minimum distance between the two parts. A predicted contact event is
a contiguous frame run of at least $\delta_{\min} = 4$ frames
($\approx 0.13\,\text{s}$ at $30\,\text{Hz}$) on which this distance stays below
$\epsilon$. A predicted event matches a GT event if
\emph{(a)} both body-part identities agree strictly (no left/right
or hand/arm equivalence; the GT body-part identity is the
skinning-argmax label of $v_e^{\text{act}}, v_e^{\text{par}}$) and
\emph{(b)} the predicted and GT time windows overlap (Jaccard-style
$\geq\!1$ shared frame; we report identical conclusions under
$\tau_{0.5}$ and $\text{IoU}_{0.3}$ overlap variants). Per-clip
Precision is the fraction of predicted events with at least one
match; clips with no predicted events are excluded.

\paragraph{Why the two metrics use different granularities}
Recall verifies \emph{known} GT events, each of which already names the exact
actor and partner vertices, so it can use the tightest per-vertex test. Precision
must instead \emph{discover unknown} predicted contacts and penalise spurious ones,
which means scanning the whole retargeted motion for emergent G1--partner
proximities. A per-vertex scan is ill-posed here because the actor side is the
G1 surface, a different embodiment, and exhaustive vertex-vs-vertex testing is
dominated by mesh-resolution noise, so we score Precision at the coarser body-part
level. We did try a vertex-pair precision and discarded it. Scoring only the
GT-named pairs rewards conservative methods that attempt few contacts, which earns
high precision by abstention, and never counts the spurious near-contacts a method
generates elsewhere. By contrast, the VLM contact-label evaluation of
Sec.~\ref{sec:supp:contact-annotation} is a \emph{label-vs-label} setting, with
both sides annotations on the same SMPL-X mesh, so it uses the per-vertex test
for both precision and recall.

\paragraph{Jitter and Penetration} \emph{Jitter} measures motion smoothness as
the mean magnitude of the third time derivative (jerk) of the humanoid
body-keypoint trajectories. With $\mathbf{p}_{t,k}$ the world position of G1
body keypoint $k$ at frame $t$, taken over the $23$ link origins including the
root, and $\Delta t = 1/\text{fps}$ the clip timestep,
\begin{equation}
\dddot{\mathbf{p}}_{t,k} = \frac{\mathbf{p}_{t+1,k} - 3\,\mathbf{p}_{t,k} + 3\,\mathbf{p}_{t-1,k} - \mathbf{p}_{t-2,k}}{\Delta t^{3}},
\qquad
\mathrm{Jitter} = \operatorname*{mean}_{t,k}\,\big\|\dddot{\mathbf{p}}_{t,k}\big\|,
\end{equation}
averaged over interior frames and keypoints within a clip and then over clips,
in $\text{m}/\text{s}^3$, where lower is smoother. \emph{Penetration} measures how
far the humanoid interpenetrates the partner's body. For each frame we sample
surface points $\mathbf{x}_{i,t}$ on the contact-relevant G1 links and compute,
directly against the partner SMPL-X body mesh, the depth $\delta_{i,t}$ by which
each point lies inside it, with $\delta_{i,t} = 0$ for points outside the mesh.
The per-frame penetration is the deepest such depth,
$\mathrm{pen}_t = \max_i \delta_{i,t}$, which we average over frames within a clip
and then over clips and report in centimetres, where lower is better.
Penetration of the humanoid with itself is not counted.

\paragraph{Pool definitions}
Jitter and Penetration are reported on the full $N{=}1100$
valid-sequence pool, the VLM-label-filtered Inter-X corpus defined in
Sec.~\ref{sec:supp:data}, of which the $80$ GT-annotated clips are a
subset. OmniRetarget is scored on $N{=}766$ of these, because its hard
collision solver drops the remaining ${\approx}30\%$ of the clips as infeasible. Recall and
Precision are reported on the $80$ GT-annotated clips only.

\begin{table}[t]
  \centering
  \small
  \caption{\textbf{Retargeting Recall vs.\ contact threshold $\epsilon$ (cm)} (per-event, across-clip mean, \%). The $\epsilon{=}10$ column reproduces the Recall column of the main paper's retargeting table. Relaxing $\epsilon$ shows that for Ours the contact is in fact established but slightly displaced: Recall rises from $59.4\%$ at $10\,\text{cm}$ to $80.0\%$ at $15\,\text{cm}$ and $92.0\%$ at $20\,\text{cm}$.}
  \label{tab:supp:retarget-sweep}
  \setlength{\tabcolsep}{5pt}
  \begin{tabular}{lccccccc}
    \toprule
    Method $\backslash$ $\epsilon$ (cm) & $3$ & $5$ & $10$ & $15$ & $20$ & $25$ & $30$ \\
    \midrule
    GMR                 & 0.0  & 0.0  & 0.4  & 0.6  & 1.3  & 4.9  & 10.6 \\
    OmniRetarget        & 2.7  & 4.9  & 22.1 & 43.3 & 58.4 & 77.3 & 86.2 \\
    \midrule
    Ours w/o Stage 2 co-opt & 1.7  & 5.0  & 13.4 & 33.6 & 54.4 & 72.4 & 79.4 \\
    Ours w/o VLM filtering & \textbf{12.1} & 23.1 & 58.0 & 79.3 & 87.2 & 91.3 & \textbf{97.2} \\
    \rowcolor{LightGreen}
    Ours                & \textbf{12.1} & \textbf{27.1} & \textbf{59.4} & \textbf{80.0} & \textbf{92.0} & \textbf{94.3} & 95.9 \\
    \bottomrule
  \end{tabular}
\end{table}

\begin{table*}[t]
  \centering
  \small
  \caption{\textbf{Per-action retargeting Recall and Precision} ($\epsilon{=}10\,\text{cm}$, per-event, across-clip mean$_{\pm\text{std}}$, \%). For each action the two columns are Recall (R, vertex-pair-targeted) and Precision (P, body-part $\tau_0$); the Total columns equal the Recall and Precision columns of the main paper's retargeting table. Recall uses $16$ GT clips per action ($58$ in total for OmniRetarget, whose collision solver makes the rest infeasible); Precision averages over clips with at least one predicted event.}
  \label{tab:supp:retarget-peraction}
  \setlength{\tabcolsep}{3pt}
  \begin{tabular}{l cc cc cc cc cc cc}
    \toprule
    & \multicolumn{2}{c}{Hug} & \multicolumn{2}{c}{Hand-shake} & \multicolumn{2}{c}{Pat-on-back} & \multicolumn{2}{c}{High-five} & \multicolumn{2}{c}{Link-arms} & \multicolumn{2}{c}{Total} \\
    \cmidrule(lr){2-3}\cmidrule(lr){4-5}\cmidrule(lr){6-7}\cmidrule(lr){8-9}\cmidrule(lr){10-11}\cmidrule(lr){12-13}
    Method & R & P & R & P & R & P & R & P & R & P & R & P \\
    \midrule
    GMR                     & $0.0$  & $1.3$  & $0.0$  & $0.0$  & $0.0$  & $7.6$ & $0.0$  & $0.0$  & $2.1$  & $1.9$ & $0.4$  & $2.3$ \\
    OmniRetarget            & $17.4$ & $4.5$  & $20.8$ & $\mathbf{21.8}$ & $16.7$ & $3.4$ & $9.4$  & $24.5$ & $54.2$ & $7.6$ & $22.1$ & $12.5$ \\
    \midrule
    Ours w/o Stage 2 co-opt & $17.5$ & $5.2$  & $0.0$  & $19.6$ & $9.4$  & $\mathbf{8.2}$ & $6.2$  & $24.1$ & $33.9$ & $\mathbf{9.7}$ & $13.4$ & $11.8$ \\
    Ours w/o VLM filtering  & $\mathbf{38.1}$ & $6.4$  & $\mathbf{84.4}$ & $15.3$ & $42.4$ & $5.0$ & $49.0$ & $27.8$ & $\mathbf{75.0}$ & $4.5$ & $58.0$ & $11.6$ \\
    \rowcolor{LightGreen}
    Ours                    & $31.7$ & $\mathbf{6.8}$  & $81.2$ & $19.7$ & $\mathbf{42.9}$ & $5.9$ & $\mathbf{70.8}$ & $\mathbf{33.3}$ & $68.8$ & $5.2$ & $\mathbf{59.4}$ & $\mathbf{14.2}$ \\
    \bottomrule
  \end{tabular}
\end{table*}

\paragraph{Reading the per-action results} Table~\ref{tab:supp:retarget-peraction} places
Recall and Precision side by side per action, which separates two failure modes that the
aggregate numbers hide. Stage~2 co-optimisation is what creates contact at all. Without it
Recall averages $13.4\%$ and is $0\%$ on hand-shake, because the Stage-1 initialisation never
closes the palm-to-palm gap. Given Stage~2, the VLM filter helps most where contact is short
and palm-localised. On high-five it lifts both axes (R $49.0\!\to\!70.8$, P
$27.8\!\to\!33.3$), the best case for our method, whereas on hand-shake the distance
candidates already localise the grip, so the filter trades a little Recall ($84.4\!\to\!81.2$)
for cleaner Precision. The actions where $F_1$ is lowest split by \emph{which} axis is binding.
Pat-on-back is recall-rich but precision-poor (close-range arm motion fires many spurious
near-contacts), while hug is recall-bound for every method, because $33\%$ of its GT events put
the actor on the torso/pelvis, outside the hand/forearm/upper-arm regions the predictor
enumerates, a coverage limit of the candidate generator, not of the retargeter.

\paragraph{Reading the absolute values and spread} Both metrics are deliberately strict and
are \emph{not} binary interaction-success rates. Recall demands the registered actor point fall
within $\epsilon{=}10\,\text{cm}$ of the \emph{exact} GT partner vertex. Relaxing $\epsilon$ shows the
contact is present but slightly displaced (Ours $59.4\%\!\to\!80.0\%\!\to\!92.0\%$ at
$10/15/20\,\text{cm}$, Table~\ref{tab:supp:retarget-sweep}). Precision counts every body-part near-pass
as a predicted event, so a method that genuinely reaches contact sweeps past many partner parts
during approach and retraction, emitting $\approx\!9\times$ more predicted events than GT
($1740$ vs.\ $203$ for Ours) and capping Precision near $14\%$. The large across-clip std in
the main paper's retargeting Recall (Ours $59.43_{\pm38.91}$) is a property of this binary
per-event metric on short clips. Per-clip Recall averages only $2$--$3$ events and so takes
coarse values in $\{0,\tfrac12,\tfrac23,1\}$, whereas the action-level means span a far narrower
$32$--$81\%$, i.e.\ the spread is per-clip difficulty, not systematic action differences.
Finally, OmniRetarget is scored on only its $58$ feasible clips, which \emph{favours} it at
larger thresholds (unreachable clips are dropped, not counted as misses). The $\epsilon{=}10\,\text{cm}$
comparison where Ours leads on both axes is therefore the conservative reading. What these
metrics do \emph{not} capture, namely whether the interaction visually works, is addressed by the
user study (Table~\ref{tab:supp-userstudy}) and the real-robot results (Sec.~\ref{sec:real-deploy}).

\subsection{Retargeting User Study: protocol and per-question results}
\label{sec:supp:userstudy}

\begin{figure}[t]
\centering
\includegraphics[width=0.65\linewidth]{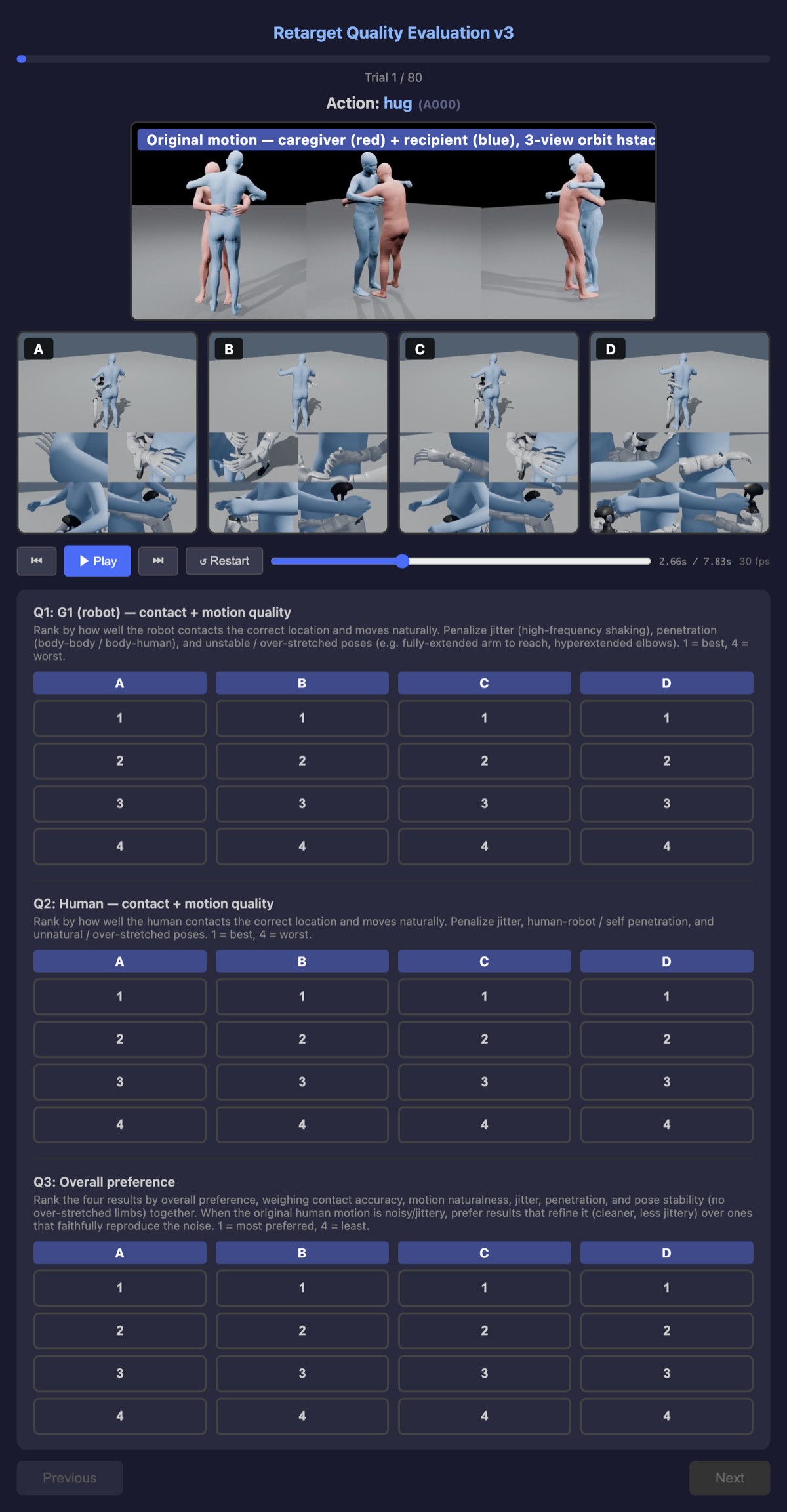}
\caption{\textbf{User-study web interface (one trial).} The original
demonstration ($3$-view orbit, top) and the four blinded retarget conditions
A--D (middle) are played synchronously; the participant gives a strict
$1$--$4$ ranking for each of the three questions (bottom). Full ranking
criteria are in Sec.~\ref{sec:supp:userstudy}.}
\label{fig:supp:userstudy-ui}
\end{figure}

\paragraph{Sequences and conditions} The study set consists of
$80$ human--human interaction sequences ($16$ per
action $\times\, 5$ actions: \emph{Hug}, \emph{Hand-shake},
\emph{Pat-on-back}, \emph{High-five}, \emph{Link-arms}). For each
sequence we render the four retarget conditions of
the main paper (GMR, OmniRetarget, Ours w/o VLM filtering,
Ours), all displayed side-by-side in randomised positions
$\{1,2,3,4\}$. Above the four method panels we show the original
SMPL-X demonstration as a $3$-view orbit reference. Both the
sequence order and the position-to-method assignment are
per-participant random but deterministic from a hash of the
participant's name, so that interrupted sessions resume on the same
trial layout.

\paragraph{Task} For every trial the participant assigns a strict
ranking over the four conditions, $\{1, 2, 3, 4\}$ with $1$ = best
and $4$ = worst (permutation, no ties), separately for three
independent questions: (Q1) \emph{G1 contact $+$ motion quality},
(Q2) \emph{Human contact $+$ motion quality}, (Q3) \emph{Overall
preference}. Participants were instructed to weight five criteria together when
forming each ranking. \emph{(i) Pose} is the consistency of the rendered
body pose with the human demo above. \emph{(ii) Contact} is whether
contact occurs at the location implied by the action label, for example
torso-to-torso for hug and hand-to-hand for hand-shake. \emph{(iii) Jitter}
is the absence of high-frequency shaking. \emph{(iv) Penetration} is the
magnitude of mesh interpenetration with the partner. \emph{(v) Stability}
is the absence of over-stretched limbs and unnatural joint bends. They were
also told to prefer a refined-but-cleaner result over a faithful
reproduction of any noise present in the original demonstration. The
study is delivered as a web app that records each
(participant, sequence, question, condition, rank) row to a
serverless key-value store on submission of every trial.

\paragraph{Results} We collected responses from $40$ participants and
restrict the analysis to the $N{=}28$ who finished all $80$ trials
with valid $\{1,2,3,4\}$ permutations on all three questions. For
each participant we compute the mean rank of each condition averaged
over the $80$ sequences; Table~\ref{tab:supp-userstudy} reports the
across-participant mean $\pm$ standard error (SEM, $n{=}28$) per
(question, condition) pair.

\begin{table}[t]
  \centering
  \small
  \caption{\textbf{User study: per-question mean ranks $\pm$ SEM
  ($n{=}28$ participants, $80$ sequences each, lower is better).}
  Each entry is the across-participant mean of per-participant
  per-condition mean ranks. Ranks lie in $\{1,2,3,4\}$ with
  $1{=}\text{best}$.}
  \label{tab:supp-userstudy}
  \begin{tabular}{lccc}
    \toprule
    Method & Q1: G1 contact & Q2: Human contact & Q3: Overall \\
           & + motion\,$\downarrow$ & + motion\,$\downarrow$ & preference\,$\downarrow$ \\
    \midrule
    GMR                                 & $3.52 \pm 0.07$ & $3.45 \pm 0.07$ & $3.61 \pm 0.04$ \\
    OmniRetarget                        & $2.64 \pm 0.04$ & $2.49 \pm 0.05$ & $2.57 \pm 0.04$ \\
    \midrule
    Ours w/o VLM filtering              & $2.05 \pm 0.04$ & $2.18 \pm 0.05$ & $2.06 \pm 0.03$ \\
    \rowcolor{LightGreen}
    Ours                                & $\textbf{1.79} \pm 0.04$ & $\textbf{1.88} \pm 0.05$ & $\textbf{1.76} \pm 0.03$ \\
    \bottomrule
  \end{tabular}
\end{table}

The full pipeline takes the best mean rank on every question. The
largest gap on Q1 (G1 contact: $1.79$ vs.\ OmniRetarget's $2.64$, a
$0.85$-rank improvement) confirms that the VLM-guided contact annotation
is the component most directly tied to robot-side contact correctness.
Removing VLM filtering (\emph{Ours w/o VLM filtering}) costs about $0.26$
rank on Q1 and $0.30$ on Q3, consistent with the precision drop reported
in the main paper's retargeting comparison. GMR is ranked last on every question across
participants ($\geq 3.45$), reflecting its lack of any contact-aware
objective.

\paragraph{Statistic and significance} Each tabulated value is an across-participant mean of
per-participant per-condition mean ranks; the reported $\pm$ is the standard error of that mean
over $n{=}28$ participants (SEM, not the across-participant std). The per-condition SEMs,
at most $0.07$, are far smaller than the gaps between adjacent methods, at least $0.26$
rank, so the orderings are unambiguous.

\section{Teacher Policy: observation breakdown, reward weights, and training}
\label{sec:supp:teacher}

\paragraph{Observation} The teacher's proprioception vector $\mathbf{o}^{\text{prop}}_t = [\boldsymbol{\omega}_b,\, \mathbf{q},\, \dot{\mathbf{q}},\, \mathbf{a}^{\text{joint}}_{t-1}]$ concatenates base angular velocity (3), 29 joint positions, 29 joint velocities, and the previous joint-level action (29) output by the HOMIE conversion below. The reference and partner inputs comprise the target G1 joint positions and velocities at the current frame and at four future offsets ($[1,2,4,8]$ steps), the reference root orientation in the base frame, and the partner's root and body positions relative to the robot base. Small per-term observation noise (joint positions $\sigma{=}0.01$, joint velocities $\sigma{=}0.5$, base angular velocity $\sigma{=}0.2$, partner positions $\sigma{=}0.1$) regularises the policy during rollouts. The critic is noise-free and additionally observes privileged terms unavailable to the actor, namely base linear velocity, the robot's body positions and orientations in the base frame, and the reference body position.

\paragraph{Action} The teacher emits $\mathbf{a}_t \in \mathbb{R}^{18}$: four HOMIE locomotion commands $(v_x, v_y, \omega_z, h)$ and 14 residual upper-body joint offsets. The HOMIE tracker converts the 4-d command into 15 lower-body joint actions, and the 14 residuals are added in action space to the demonstration's reference arm pose. The tracker is kept frozen throughout teacher training.

\paragraph{Reward weights} The aggregate per-step reward is
\[
  r_t = 3\,r^{\text{dir}}_t + 5\,r^{\text{dist}}_t + 0.5\,r^{\text{force}}_t + 0.5\,r^{\text{root\,pos}}_t + 0.5\,r^{\text{root\,ori}}_t + r^{\text{body\,pos}}_t + r^{\text{body\,ori}}_t + r^{\text{body\,vel}}_t + r^{\text{body\,ang}}_t + r^{\text{no\,fly}}_t + r^{\text{reg}}_t,
\]
where each motion-tracking term has the exponential form $\exp(-e/\sigma^2)$, with $e$ the per-body mean squared tracking error and $\sigma$ the error scale at which the reward decays to $1/e$ (a smaller $\sigma$ enforces tighter tracking). The relative-frame body terms use body-position $\sigma = 0.3\,\text{m}$, body-orientation $\sigma = 0.4\,\text{rad}$, body-velocity $\sigma = 1.0\,\text{m/s}$, and body-angular-velocity $\sigma = 3.14\,\text{rad/s}$. Regularisation: action rate $-3{\times}10^{-2}$, joint velocity $-2{\times}10^{-4}$, joint acceleration $-1{\times}10^{-7}$, torque $-1{\times}10^{-4}$, DOF-position soft limit $-100$ (with $0.9$ soft-limit ratio), DOF-velocity soft limit $-1{\times}10^{-3}$, torque soft limit $-3{\times}10^{-2}$. The no-fly term rewards at least one foot in contact ($+1$ when satisfied).

\paragraph{Initialisation} Episodes use a hybrid \textit{physical state initialisation}: with probability $0.5$ the episode restarts from the policy's own recent on-policy state; otherwise it is seeded from the demonstration's first frame perturbed by Gaussian noise with joint positions $\sigma = 0.1$, root translation $\sigma_{xyz} = (0.05, 0.05, 0.01)\,\text{m}$, root orientation $\sigma = (0.1, 0.1, 0.2)\,\text{rad}$, root linear velocity $\sigma = (0.5, 0.5, 0.2)\,\text{m/s}$, and root angular velocity $\sigma = (0.52, 0.52, 0.78)\,\text{rad/s}$. The initial action noise std is $\sigma_{\text{init}} = 0.5$. The partner is driven by a reference PD controller that tracks the demonstration.

\paragraph{PPO} Both actor and critic are 5-layer MLPs with hidden sizes $[512, 512, 512, 256, 128]$ and ELU activations. We use PPO~\cite{schulman2017proximal} with $\gamma = 0.99$, GAE $\lambda = 0.95$, clipping $0.2$, 8 update epochs per rollout, 4 minibatches, entropy coefficient $0.005$, and AdamW~\cite{loshchilov2018decoupled} with learning rate $10^{-5}$ for both actor and critic. We collect 24 environment steps per iteration across 8{,}192 parallel Isaac Sim environments on a single L40S GPU.

\paragraph{Evaluation metrics: MPJPE and Contact Recall} The teacher is
evaluated with MPJPE and Contact Recall, and the student that distils it with
Contact Recall. \emph{MPJPE} (mean
per-joint position error) measures tracking fidelity, the mean Euclidean distance
between the humanoid body keypoints under physics and the same keypoints of the
retargeted reference the policy tracks,
\begin{equation}
\mathrm{MPJPE} = \operatorname*{mean}_{t,k}\,\big\|\mathbf{p}^{\text{sim}}_{t,k} - \mathbf{p}^{\text{ref}}_{t,k}\big\|,
\end{equation}
reported in centimetres. \emph{Contact Recall} measures how often the policy
realises the demonstrated contacts under physics. Manual annotation does not
scale to the policy evaluation pools (all student-pool motions for the teacher, the held-out
split for the student), so we use the VLM-verified contact
annotations of Sec.~\ref{sec:supp:contact-annotation} as ground truth. For every
annotated contact frame we carry the prescribed actor contact site onto the G1
surface through the registration of Sec.~\ref{sec:supp:registration}, measure its
distance to the annotated partner vertex under partner forward kinematics, and
count the frame as satisfied when this distance is below $15\,\text{cm}$ for the
teacher and $30\,\text{cm}$ for the student. Contact
Recall is the fraction of satisfied (frame, contact site) pairs, pooled over all
evaluated sequences.

\subsection{Contact Reward: coefficients and force gate}
\label{sec:supp:contact-reward}

Each VLM-verified annotation specifies, per frame, the active contact pairs between fixed G1 contact links and partner surface points. The six candidate links are the left and right wrist, elbow, and shoulder. For a pair $h$ active at time $t$, let $\mathbf{p}^h_t$ denote the G1 contact point, $\mathbf{j}^h_t$ the target partner vertex, $\mathbf{n}^h_t$ the partner surface normal at that vertex, and $F^h_t$ the contact-force magnitude on the corresponding G1 link. The G1 contact point is recovered by forward kinematics from the mesh-projection registration of Sec.~\ref{sec:supp:registration}, $\mathbf{p}^h_t = T_{\text{link}^h}(\mathbf{q}_t)\,\mathbf{p}_{\text{local}^h}$, where $(\text{link}^h, \mathbf{p}_{\text{local}^h})$ is the registered G1 link and surface point for the contact. The world-frame target $\mathbf{j}^h_t$ is the partner contact-vertex position from the SMPL-X forward kinematics of the reference motion.

We supervise the teacher with three complementary contact rewards. The first drives the demonstrated contact closed and gates it on real force,
\begin{equation}
  r^{\text{dist}}_t = \Big\langle\, \phi(F^h_t)\,\exp\!\big(-\sigma_d\,|\mathbf{p}^h_t-\mathbf{j}^h_t|^2\big) \Big\rangle_h,
  \qquad
  \phi(F) = \min\!\big(\exp(F-\tau),\, 1\big),
  \label{eq:contact-reward}
\end{equation}
averaged over the active pairs $h$, with sharpness $\sigma_d = 50\,\text{m}^{-2}$ and force-gate threshold $\tau = 1\,\text{N}$ (the gate is active because robot--partner collisions are enabled). The second encourages the contact to be approached from the physically correct side and discourages satisfying the distance objective through interpenetration,
\begin{equation}
  r^{\text{dir}}_t = \Big\langle\, \mathrm{clamp}\!\big(\hat{\mathbf{v}}^h_t \cdot \mathbf{n}^h_t,\, 0,\, 1\big)\,\cdot\,\mathrm{clamp}\!\big(\mathbf{n}^{h,\text{G1}}_t \cdot (-\hat{\mathbf{v}}^h_t),\, 0,\, 1\big) \Big\rangle_h,
  \qquad
  \hat{\mathbf{v}}^h_t = \frac{\mathbf{p}^h_t-\mathbf{j}^h_t}{\max\!\big(|\mathbf{p}^h_t-\mathbf{j}^h_t|,\, \epsilon\big)},
  \label{eq:contact-dir}
\end{equation}
with $\epsilon = 0.02\,\text{m}$. The first factor requires the approach direction to align with the partner surface normal $\mathbf{n}^h_t$, and the second requires the robot's own contact-surface normal $\mathbf{n}^{h,\text{G1}}_t$ to face the partner. The third regulates the established contact force. Per pair, with contact label $l^h_t$ and force band $[F_{\min}, F_{\max}] = [5, 50]\,\text{N}$,
\begin{equation}
  r^{\text{force},h}_t =
  \begin{cases}
    1, & l^h_t = 1,\ F^h_t \in [F_{\min}, F_{\max}], \\
    \exp\!\big(-c_u (F_{\min}-F^h_t)\big), & l^h_t = 1,\ F^h_t < F_{\min}, \\
    \exp\!\big(-c_o (F^h_t-F_{\max})\big), & l^h_t = 1,\ F^h_t > F_{\max}, \\
    -c_w F^h_t, & l^h_t = 0,\ F^h_t > F_{\min}, \\
    0, & \text{otherwise},
  \end{cases}
  \label{eq:contact-force}
\end{equation}
summed over pairs to give $r^{\text{force}}_t = \sum_h r^{\text{force},h}_t$, with $c_u = 0.5$, $c_o = 0.05$, and $c_w = 0.005$.

In the aggregate teacher reward these terms are weighted $3.0\,r^{\text{dir}}_t + 5.0\,r^{\text{dist}}_t + 0.5\,r^{\text{force}}_t$.

\section{Limitations}
\label{sec:supp:limitations}

\paragraph{Fixed-view, passive perception} The student observes a single forward-facing depth camera and does not control where it looks. Under heavy self-occlusion the partner can leave the field of view entirely, and the policy cannot actively reposition the camera to recover the missing state. Coupling the interaction policy with active head or gaze control, so the robot moves to keep the relevant contact region in view, is a natural extension.

\paragraph{Limited dexterity} REACT drives the body through HOMIE locomotion commands plus residual upper-body joint offsets and treats the hands as rigid end effectors. It does not actuate fingers, so interactions that require grasping, fine in-hand manipulation, or force-sensitive hand shaping, such as a firm but gentle hand-hold or adjusting a grip, are out of scope. Extending the action space to dexterous hands is needed for this class of contact.

\paragraph{Simple interaction repertoire} We evaluate on five short, largely upper-body interactions (hug, high-five, handshake, link-arms, pat-on-back). More complex and longer-horizon behaviors that involve sustained multi-contact load bearing and tighter coupling to the partner's intent are beyond what the current contact reward and demonstration pipeline target, and scaling to them is left to future work.

\section{VLM Verification Prompt (verbatim)}
\label{sec:supp:vlm-prompt}
The template below is the exact text sent to Gemini~2.0 Flash for each $3\,\text{s}$ video
chunk, alongside the horizontally concatenated three-view frames (extracted at $5\,\text{fps}$ and
cropped to the two figures). The prompt first asks the model to reason about the contact
patterns expected for the action, then to classify each distance-based candidate (mocap
proximity $<15$\,cm) into one of \texttt{essential}, \texttt{incidental},
\texttt{unnecessary}, or \texttt{avoid}, emitting a structured JSON verdict per candidate.
Runtime fields are substituted at call time: \texttt{\{action\_name\}},
\texttt{\{timestamp\_list\}}, \texttt{\{num\_views\}} ($=3$), \texttt{\{view\_labels\}}
(\texttt{view-0, view-1, view-2}), \texttt{\{text\_description\}} (the Inter-X interaction
description), \texttt{\{candidates\_formatted\}} (the enumerated distance candidates), and
\texttt{\{num\_candidates\}}. Unicode arrows in the original ($\leftrightarrow$,
$\rightarrow$) are transcribed here as \texttt{<->} and \texttt{->}.

\begin{promptbox}
\begin{Verbatim}[fontsize=\scriptsize]
You are an expert analyst of close human-human interactions. Your task
is to classify distance-based contact candidates by their **semantic
role** relative to the action being performed.

## Video Information
- Action type: {action_name}
- Frame timestamps: {timestamp_list}
- **Multi-view**: Each frame image contains {num_views} views
  concatenated horizontally, labeled as {view_labels} (from left to
  right)

## Person Identification
- **P1 (Blue person)**: The person rendered in BLUE color
- **P2 (Orange person)**: The person rendered in ORANGE color

## Interaction Description
{text_description}

NOTE: "first person" / "second person" in the description may NOT
correspond to P1(Blue)/P2(Orange). Use the semantic content to
understand WHAT contact should occur, then visually identify WHO is
performing it.

## Body Part Naming Convention
- `left_wrist` / `right_wrist`: Represents the entire hand including all
  fingers
- `left_elbow` / `right_elbow`: Represents the forearm region
- `left_shoulder` / `right_shoulder`: Represents the upper arm and
  shoulder region
- Contact categories for the initiating part: "hand", "forearm",
  "upper_arm"

## Contact Surface Definitions

**CONTACTING SURFACE** (the person initiating contact):
- `"palm/inner"`: Palm side of hand, inner side of arm (gripping/
  embracing side)
- `"back/outer"`: Back of hand, outer side of arm

**CONTACTED SURFACE** (the person being touched):
- `"inner"`: Inner/front surface of the body part being touched
- `"outer"`: Outer/back surface of the body part being touched

## ============================================================
## STEP 1: Determine Expected Contact Patterns
## ============================================================

Before evaluating candidates, you MUST FIRST reason about what joint
contact pairs and surfaces are semantically expected for this action.

**Reference patterns by action type** (use as guidance - adapt based on
the text description):

### High-five
- **essential**: hand(palm/inner) <-> wrist(palm/inner) - palm-to-palm
  contact
- **avoid**: hand <-> head (hand hitting head)
- **NOT expected** (mark as `unnecessary`): hand <-> elbow, forearm <->
  wrist, upper_arm <-> elbow - these are incidental proximity from arm
  extension, NOT actual contact in a high-five

### Hug
- **essential**: hand(palm/inner) <-> spine3/back(outer) - palms
  embracing the back
- **essential**: forearm(inner) <-> shoulder/collar(outer) - arms
  wrapping around shoulders
- **incidental**: upper_arm(inner) <-> shoulder/upper_arm(outer) - may
  occur but not the defining contact
- **incidental**: hand(palm/inner) <-> shoulder(outer) - hand resting on
  shoulder during hug
- **avoid**: hand <-> head, elbow <-> face

### Pat on back
- **essential**: hand(palm/inner) <-> spine3/back(outer) - palm patting
  the back
- **incidental**: hand(palm/inner) <-> shoulder/collar(outer) - patting
  the shoulder area
- **avoid**: hand <-> head
- **NOT expected** (mark as `unnecessary`): forearm <-> wrist, upper_arm
  <-> elbow

### Link arms
- **essential**: forearm(inner) <-> forearm(inner) - inner forearms
  interlocked
- **essential**: hand(palm/inner) <-> elbow/forearm(outer) - hand
  gripping the other's elbow
- **incidental**: upper_arm <-> forearm - may touch during interlocking
- **NOT expected** (mark as `unnecessary`): hand <-> spine3, hand <->
  neck

### Walk towards
- Generally NO intentional contact expected. Mark candidates as
  `unnecessary` unless the text description explicitly mentions touching
  (e.g., handshake, bump).

Based on action type "{action_name}" and the text description, determine
the specific expected contact patterns including sides (left/right) and
surfaces.

## ============================================================
## STEP 2: Classify Each Candidate
## ============================================================

The following contact candidates were detected based on spatial
proximity (distance < 15cm):

{candidates_formatted}

**CRITICAL PRINCIPLE**: Spatial proximity alone does NOT mean the
contact is semantically meaningful. During any action, many body parts
come close together, but only specific joint pairs represent actual
intended contact. Your job is to classify each candidate's semantic
role.

### Semantic Labels

Each candidate MUST be assigned one of these labels:

| semantic_label | is_valid | Meaning |
|---|---|---|
| `essential` | true | Required for the action to occur (e.g.,
  palm-to-palm in high-five) |
| `incidental` | true | May naturally occur but is not required (e.g.,
  upper arms touching during a hug) |
| `unnecessary` | false | Not part of the action - incidental proximity
  or mocap artifact (e.g., hand near elbow during high-five) |
| `avoid` | false | Contact that should NOT occur - collision or
  unnatural contact (e.g., hand hitting head, body penetration) |

### Classification Criteria

For each candidate, evaluate:
1. **Semantic match**: Does the joint pair match an expected contact
   pattern from Step 1?
2. **Surface match**: Are the contacting/contacted surfaces consistent
   with the expected pattern?
3. **Visual evidence**: Examine ALL {num_views} views. Is the contact
   visible in at least one view with HIGH confidence?
4. **Duration**: Is the duration appropriate for the action type?
   - Hug: 2-10s typical. <0.5s too brief.
   - High-five: 0.3-2s acceptable.
   - Pat on back: 0.3-1s per pat, series may span longer.
   - Link arms: 2-10s+ sustained.
5. **Temporal context**: If multiple candidates overlap in time, they
   may represent the same physical event. Mark the most semantically
   appropriate one as essential; mark redundant ones as incidental or
   unnecessary.

### is_valid rules:
- `essential` -> is_valid = true
- `incidental` -> is_valid = true
- `unnecessary` -> is_valid = false
- `avoid` -> is_valid = false

### Multi-view validation:
- For `essential` or `incidental`: at least one view MUST show the
  contact with HIGH confidence.
- For `unnecessary` or `avoid`: evidence_views should be empty.

CRITICAL: You MUST evaluate ALL {num_candidates} candidates listed above.

Output ONLY valid JSON in this format:
```json
{
  "expected_contacts": [
    {
      "pattern": "hand <-> wrist",
      "semantic_label": "essential",
      "contacting_surface": "palm/inner",
      "contacted_surface": "palm/inner",
      "description": "High-five: palm-to-palm contact between right hands"
    },
    {
      "pattern": "hand <-> head",
      "semantic_label": "avoid",
      "contacting_surface": "",
      "contacted_surface": "",
      "description": "Hand should not hit the head"
    }
  ],
  "validations": [
    {
      "candidate_id": 0,
      "is_valid": true,
      "semantic_label": "essential",
      "matched_pattern": "hand <-> wrist",
      "confidence": "high",
      "evidence_views": [0, 2],
      "contacting_surface": "palm/inner",
      "contacted_surface": "palm/inner",
      "reason": "P1(Blue)'s right hand touching P2(Orange)'s right
      wrist for 0.7s. Matches expected high-five palm-to-palm pattern.
      High confidence in view-0 and view-2."
    },
    {
      "candidate_id": 1,
      "is_valid": false,
      "semantic_label": "unnecessary",
      "matched_pattern": "none",
      "confidence": "low",
      "evidence_views": [],
      "contacting_surface": "",
      "contacted_surface": "",
      "reason": "P1(Blue)'s right hand near P2(Orange)'s right elbow.
      Does NOT match any expected high-five pattern. Hand-to-elbow
      proximity is incidental to arm extension, not an actual contact."
    },
    {
      "candidate_id": 2,
      "is_valid": false,
      "semantic_label": "avoid",
      "matched_pattern": "hand <-> head",
      "confidence": "high",
      "evidence_views": [1],
      "contacting_surface": "",
      "contacted_surface": "",
      "reason": "P2(Orange)'s hand appears to collide with P1(Blue)'s
      head. This is an undesirable contact that should be avoided."
    }
  ]
}
```

RULES:
- Output `expected_contacts` array FIRST, listing all expected patterns
  for this action (with semantic_label and surfaces)
- Provide a validation entry for EVERY candidate ({num_candidates}
  total)
- Every validation MUST include `semantic_label` (essential/incidental/
  unnecessary/avoid) and `matched_pattern`
- `matched_pattern`: which expected pattern this candidate matches, or
  `"none"` if no match
- **is_valid=true REQUIRES**: semantic_label is essential or incidental,
  AND at least one view with HIGH confidence
- **is_valid=false**: if semantic_label is unnecessary or avoid,
  regardless of distance
- Use P1(Blue) and P2(Orange) terminology
- For `essential`/`incidental` contacts: ALWAYS specify
  contacting_surface and contacted_surface
- For `unnecessary`/`avoid` contacts: leave contacting_surface and
  contacted_surface as empty strings
\end{Verbatim}
\end{promptbox}

\end{document}